\documentclass[a4paper,fleqn]{cas-dc}

\usepackage[numbers]{natbib}
\usepackage{amsmath,amsfonts}
\usepackage{mathrsfs}
\usepackage{algorithmic}
\usepackage{algorithm}
\usepackage{array}
\usepackage{diagbox}
\usepackage{booktabs}
\usepackage{bm}
\usepackage{textcomp}
\usepackage{url}
\usepackage{verbatim}
\usepackage{threeparttable}
\usepackage{multirow}
\usepackage{textcomp}
\usepackage{balance}
\usepackage{pifont}
\usepackage{color}
\usepackage{amsthm}
\usepackage{longtable}
\usepackage{array}
\usepackage{caption}
\usepackage{subcaption}
\usepackage{graphicx}
\usepackage{float}
\usepackage[dvipsnames]{xcolor}
\definecolor{ccr}{RGB}{10,110,150} 
\usepackage{hyperref}
\hypersetup{citecolor=ccr}

\def\tsc#1{\csdef{#1}{\textsc{\lowercase{#1}}\xspace}}
\tsc{WGM}
\tsc{QE}
\begin{document}
\let\WriteBookmarks\relax
\def\floatpagepagefraction{1}
\def\textpagefraction{.001}
%\let\printorcid\relax % 可去掉页面下方的ORCID(s)

% \shorttitle{<short title of the paper for running head>} 
%\shorttitle{Scene-Aware Vectorized Memory Multi-Agent Framework with Cross-Modal Differentiated Quantization VLMs for Visually Impaired Assistance}    

% \shortauthors{<short author list for running head>}
%\shortauthors{X. Wang et al.}

\title[mode = title]{Scene-Aware Vectorized Memory Multi-Agent Framework with Cross-Modal Differentiated Quantization VLMs for Visually Impaired Assistance}  

% \tnotemark[<tnote number>] 
\tnotemark[1]

\tnotetext[1]{This work was supported in part by the National Natural Science Foundation of China under Grants 62406062 and 62276055, in part by the Natural Science Foundation of Sichuan Province under Grant 2024NSFSC1476, in part by the Sichuan Science and Technology Program under Grant 2023YFG0288, in part by the National Science and Technology Major Project under Grant 2022ZD0116100, in part by the Sichuan Provincial Major Science and Technology Project under Grant 2024ZDZX0012,Joint Fund for Medicine-Engineering Interdisciplinary Research of UESTC and Sichuan Provincial People's Hospital under Grant ZYGX2025YGLH001.}


% \author[<aff no>]{<author name>}[<options>]

% \cormark[<corr mark no>]

% \fnmark[<footnote mark no>]

% \ead{<email address>}

% \ead[url]{<URL>}

% \credit{<Credit authorship details>}

% \affiliation[<aff no>]{organization={},
% %          citysep={}, % Uncomment if no comma needed between city and postcode

% \author[<aff no>]{<author name>}[<options>]

% Footnote of the second author

% Email id of the second author
% \ead{}



% \affiliation[<aff no>]{organization={},
% %          citysep={}, % Uncomment if no comma needed between city and postcode

% \cortext[1]{Corresponding author}



\author[1,2]{Xiangxiang Wang}%[role=Co-ordinator, suffix=Jr]
\ead{xxwang@uestc.edu.cn}

\author[1]{Xuanyu Wang}%[type=editor,
\ead{xywangmars@gmail.com} 

\author[3]{YiJia Luo}%[style=chinese]
\ead{luoyijia_hit@outlook.com} 

\author[1]{Yongbin Yu}%[style=chinese]
\fnmark[$\ast$]
\ead{ybyu@uestc.edu.cn}

\author[1]{Manping Fan}%[role=Co-ordinator, suffix=Jr]
\fnmark[$\ast$]
\ead{fmpfmp@uestc.edu.cn} 

\author[1]{Jingtao Zhang}%[role=Co-ordinator, suffix=Jr]
\ead{jtzhanguestc@gmail.com}

\author[1]{Liyong Ren}%[role=Co-ordinator, suffix=Jr]
\ead{lyren@uestc.edu.cn}

\address[1]{School of Information and Software Engineering, University of Electronic Science and Technology of China, Chengdu, Sichuan, P.R. China}

\address[2]{Sichuan Provincial Key Laboratory for Human Disease Gene Study, Sichuan Academy of Medical Sciences \& Sichuan Provincial People's Hospital, 
University of Electronic Science and Technology of China, Chengdu, Sichuan, China.}

\address[3]{Faculty of Computing, Harbin Institute of Technology, Harbin, China}

\cortext[1]{Corresponding authors}

\begin{abstract}
Visually impaired individuals face significant challenges in environmental perception. Traditional assistive technologies often lack adaptive intelligence, focusing on individual components rather than integrated systems. While Vision-Language Models (VLMs) offer a promising path to richer, integrated understanding, their deployment is severely limited by substantial computational requirements, demanding dozens of gigabytes of memory. To address these gaps in computational efficiency and integrated design, this study proposes a dual technological innovation framework: a cross-modal differentiated quantization framework for VLMs and a scene-aware vectorized memory multi-agent system. The quantization framework implements differentiated strategies, reducing memory from 38GB to 11.3GB. The multi-agent system uses vectorized memory and perception-memory-reasoning workflows to provide environmental information beyond the current view, achieving 2.83-3.52s latency to initial speech output. Experiments show the quantized 19B-parameter model only experiences a 2.05\% performance drop on MMBench and maintains 63.7 accuracy on OCR-VQA (original: 64.9), outperforming smaller models with equivalent memory. This research advances computational efficiency and assistive technology, offering comprehensive assistance in scene perception, text recognition, and navigation.
\end{abstract}


\begin{highlights}
\item Novel cross-modal quantization handles varied component sensitivity in models.
\item Scene-aware vectorized memory agents enable cross-perspective understanding.
\item Efficient dequantization and storage cuts memory needs for fast inference.
\item Flow-based multi-agent system with speech streaming offers visual aid.
\end{highlights}

% Keywords

\maketitle

\section{Introduction}

{Visually impaired individuals, representing at least 2.2 billion people globally according to WHO estimates \citep{WHO2023blindness}, face significant challenges in environmental perception and navigation \citep{okolo2024assistive}. Traditional assistive technologies often provide limited contextual understanding and lack the adaptive intelligence needed for complex, dynamic environments \citep{valipoor2024analysis}. While key components for visually impaired assistive technologies—object detection, text detection, and text-to-speech synthesis—exist, research has traditionally focused on individual components rather than integration \cite{kathiria2024assistive}. Advancing these technologies requires integrated solutions that can address users' holistic needs by providing comprehensive scene understanding.

State-of-the-art Vision-Language Models (VLMs) \citep{prajapati2024evaluation} offer a promising path to this integrated understanding. However, their practical deployment is severely limited by substantial computational resources. Models exceeding billions of parameters demand multiple high-end GPUs for inference, with memory requirements reaching dozens of gigabytes. This computational burden makes them impractical for many real-world implementations where computational resources are limited, such as on the portable devices needed for assistive technology.

Post-Training Quantization (PTQ) has shown promise for addressing these computational constraints in Large Language Models (LLMs) \citep{hu2025ostquant}, enabling model compression without extensive retraining. Methods such as GPTQ \citep{frantar2022gptq} have demonstrated impressive results. However, extending these techniques to multimodal architectures introduces unique challenges. Recent state-of-the-art methods have begun to address the distinct sensitivities of VLM components. For instance, VLMQ \citep{xue2025vlmq} identifies a modality discrepancy in tokens (i.e., limited text versus excessive and redundant vision tokens) and proposes an importance-aware objective using an enhanced Hessian matrix. Similarly, MBQ \citep{li2025mbq} also discovers a significant sensitivity difference between language and vision tokens, proposing a method to incorporate these sensitivities during the calibration process. While these advanced methods correctly identify and begin to solve discrepancies at the token-level, their focus remains primarily on data-level (token) adjustments. To the authors' best knowledge, the fundamental sensitivity differences between the architectural components themselves—where visual encoders and cross-modal processing modules have distinct computational patterns and requirements—remain underexplored.

Beyond model compression, there is a critical need for effective deployment frameworks that can translate VLM capabilities into real-world benefits for users. Multi-agent systems offer a promising approach for decomposing complex perception and reasoning tasks \citep{yang2025autohma, he2024llm}. Recent assistive systems have begun to leverage VLMs for specific, targeted applications. For example, the VLM-Drone system \citep{zhang2025vlm} employs a hierarchical structure of cloud-based VLMs and specialized hardware (a drone with TOF Laser-Camera modules) for indoor navigation and wayfinding sign interpretation. Another specialized system, NaviSense \citep{sridhar2025navisense}, focuses on open-world object retrieval by integrating VLMs with LiDAR and AR on a mobile device to provide real-time audio-haptic guidance. In the outdoor domain, other work utilizes spatial-temporal VLMs like VideoLLaMA3 to process videos for urban navigation instruction generation \citep{chao2025automated}. While these emerging solutions demonstrate VLM potential, they often (1) rely on specialized hardware such as drones or LiDAR sensors for spatial awareness, and (2) are designed for a single, specific task (e.g., object retrieval or sign-based navigation). Critically, their logic is limited to immediate perception, processing only the current field of view. To the authors' best knowledge, two key gaps persist: first, a lack of flexible, scene-aware workflows that can autonomously route between diverse tasks (e.g., text-reading, navigation, description) \citep{yang2025oc}; and second, an inability to overcome the limitations of a single perspective, as they lack a mechanism to integrate historical scene knowledge. Current agent frameworks are often too general \cite{mu2024multi} or require manual task switching, creating unnecessary cognitive burden.}

To address these {dual} challenges, a novel modular quantization framework is proposed specifically designed for VLMs, enabling effective deployment in resource-constrained environments, complemented by a flow-based {scene-aware} multi-agent assistance framework tailored to the needs of visually impaired users. The agent architecture implements a perception-memory-deliberation pipeline that continuously processes visual information, analyzes scene changes, and generates contextually appropriate assistance through natural language interaction.

The main contributions of this work are as follows:
\vspace*{-5pt}
\begin{itemize}
\setlength{\itemsep}{0pt}
\setlength{\parsep}{0pt}
\setlength{\parskip}{0pt}
 \item[(1)] An innovative cross-modal differentiated quantization framework is proposed, which addresses the varying sensitivity of different {architectural} components to quantization in vision-language models through specifically designed quantization strategies for visual encoders and cross-modal processing modules. 
  \item[(2)] A scene-aware vectorized memory multi-agent framework is designed that enables cross-perspective understanding through vector similarity retrieval, integrating historical scene knowledge to provide environmental information beyond the current view, overcoming  {immediate} single-perspective limitations for assistance.
  \item[(3)] Efficient dequantization and storage optimization techniques are developed, including bit-compressed formats and specialized computing kernels. These innovations maintain inference speed while reducing memory requirements, enabling large vision-language models to run efficiently on consumer-grade GPUs.
  \item[(4)] A flow-based multi-agent collaboration system with {low-latency}  speech streaming capability is constructed that provides comprehensive visual assistance through dedicated agents for scene classification, text recognition, obstacle detection, and environment description.
\end{itemize}

The cross-modal differentiated quantization(CMDQ) framework addresses computational efficiency challenges for VLMs, while the scene-aware vectorized memory multi-agent system enhances assistive quality for visually impaired users. These technological innovations are optimized for real-time performance on consumer-grade hardware and precise environmental understanding. The differentiated quantization strategy applies distinct processing methods to {architecturally distinct} visual encoders and cross-modal processing modules, significantly reducing memory requirements while maintaining model performance. The vectorized memory system accumulates scene knowledge through similarity-based retrieval mechanisms, compensating for {immediate} perception limitations and providing environmental information understanding beyond the current field of view. These two technologies collectively support a {computationally-efficient} and {contextually-adaptive} visual assistance system, enabling effective deployment in resource-constrained environments while providing comprehensive environmental perception support for visually impaired users.

The rest of this paper is organized as follows. Section \ref{Sec_2} introduces related work, including vision-language models, post-training quantization methods, multi-agent systems for AI applications, and retrieval-augmented generation techniques. Section \ref{Sec_3} elaborates on the proposed methods, including preliminaries, cross-modal differentiated quantization framework, and multi-agent assistance framework. Section \ref{Sec_4} evaluates the effectiveness of the proposed method through experiments, including performance comparisons on MMBench and OCR-VQA datasets, as well as practical application tests of the multi-agent assistance framework {, including RAG experiments}. {Section \ref{Sec_5} discusses the implications of the results, advantages and disadvantages of the method, and limitations of the study.} Finally, Section {\ref{Sec_6}} concludes this paper, summarizing the main contributions.

\section{Related work}\label{Sec_2}
\subsection{Vision-Language Models}
VLMs integrate visual and linguistic modalities to achieve comprehension and generation of multimodal information. In recent years, VLM architectures have evolved from independently trained components to frameworks based on LLMs. Contemporary VLMs can be categorized into three principal architectural paradigms.

Shallow alignment approaches, such as InstructBLIP \citep{li2023blip}, MiniGPT-4 \citep{zhu2023minigpt}, multimodal breast cancer prediction models \citep{vo2024frozen}, and BRAVE \citep{kar2024brave}, employ frozen pre-trained vision encoders and language models connected via Q-Formers, linear projection layers, or more complex feature consolidation mechanisms. While these methods offer implementation simplicity, their performance is limited by insufficient fusion between visual and linguistic data representations, though approaches like BRAVE attempt to mitigate this by integrating features from multiple encoders with different inductive biases.

Direct LLM training methodologies, like PaLI \citep{chen2022pali}, Qwen-VL \citep{bai2023qwen}, JM3D-LLM \citep{ji2024jm3d}, and DeepSeek-VL \citep{lu2024deepseek}, update LLM parameters directly during pre-training or supervised fine-tuning phases. Although this facilitates deeper integration, it frequently results in "catastrophic forgetting," where models trained on image-text pairs show significant degradation in purely textual task capabilities. For instance, DeepSeek-VL acknowledges this challenge by investigating "an effective VL pretraining strategy" and "carefully managing the competitive dynamics observed between vision and language modalities" to preserve LLM capabilities during multimodal training.

Vision expert-based approaches, with CogVLM \citep{wang2024cogvlm}, MM1-MoE\citep{mckinzie2024mm1}, and DeepSeek-VL2 \citep{wu2024deepseek} as prominent examples, introduce trainable vision expert modules to connect frozen pre-trained language models with image encoders. This methodology achieves the benefits of deep fusion while preserving model performance on NLP tasks.

Current study trends in VLM development indicate a shift from using independent encoders toward leveraging pre-trained LLMs as backbone networks, integrating visual information through carefully designed fusion mechanisms to achieve more effective multimodal capabilities.

\subsection{Post-Training Quantization Methods}
As Generative Pre-trained Transformer models grow in size, PTQ has emerged as an effective solution to reduce computational and storage requirements during inference. Unlike quantization during training which requires extensive retraining, PTQ can compress pre-trained models without retraining, making them more practical for billion-parameter models.

Current PTQ techniques include round-to-nearest quantization (ZeroQuant, LLM.int8()) \citep{yao2022zeroquant}, adaptive methods (AdaRound) \citep{nagel2020up}, and second-order approaches like GPTQ. Another notable approach is SmoothQuant \citep{xiao2023smoothquant}, which addresses the challenge of activation quantization by mathematically migrating quantization difficulty from activations to weights, enabling efficient 8-bit quantization with up to 1.56 times speedup and 2 times memory reduction. SEPTQ \citep{liu2025septq} offers a simplified two-step approach that calculates element-wise importance scores and performs column-wise quantization updates, showing particular strength in low-bit scenarios across models of various sizes. Meanwhile, GPTQ stands out for both its efficiency—quantizing 175B parameter models in just four GPU hours—and its effectiveness, enabling compression to 3-4 bits with minimal accuracy loss. {More recent general-purpose PTQ methods focus on mitigating activation outliers, a key challenge in low-bit quantization. For instance, AWQ \citep{lin2025awq} proposes an activation-aware method that protects salient weights by observing activation distributions, not weights. Similarly, QUAD \citep{hu2025quad} utilizes Singular Value Decomposition (SVD) to decompose and isolate activation outliers, maintaining them in full precision while quantizing the remaining components.} This allows models that previously required multiple high-end GPUs to run on a single device, with some implementations achieving up to 4.5 times inference speedups.

{Despite these advances, extending PTQ to multimodal architectures introduces unique challenges. General-purpose methods like AWQ \citep{lin2025awq} and QUAD \citep{hu2025quad}, while effective for LLMs, were not specifically designed to handle the distinct sensitivities of VLM components. Recognizing this, recent SOTA methods have begun to address VLM-specific discrepancies. For example, VLMQ \citep{xue2025vlmq} identifies a modality discrepancy in tokens (i.e., limited text versus excessive and redundant vision tokens) and proposes an importance-aware objective using an enhanced Hessian matrix. Similarly, MBQ \citep{li2025mbq} also discovers a significant sensitivity difference between language and vision tokens, proposing a method to incorporate these sensitivities during the calibration process.}

{However, while these advanced methods correctly identify and begin to solve discrepancies at the \textbf{data-level (token-level)}, their focus remains primarily on data adjustments. To the authors' best knowledge, the fundamental sensitivity differences between the \textbf{architectural components} themselves—where visual encoders and cross-modal processing modules have distinct computational patterns and requirements—remain underexplored.} This study extends GPTQ to VLMs through a novel CMDQ approach, aiming to {fill this specific architectural-level gap} and achieve efficient compression while preserving model performance across both visual and linguistic tasks.

\subsection{Multi-Agent Systems for AI Applications}
Multi-agent systems comprise autonomous computational entities that collaborate to solve complex problems exceeding individual agent capabilities. In LLM-based applications, these systems have evolved from simple interfaces to sophisticated frameworks executing complex workflows with minimal human intervention.

Agent architectures can be categorized into two primary structures. Single-agent frameworks like ReAct \citep{yao2023react} and RAISE \citep{liu2401llm} utilize planning loops and self-reflection for multi-step tasks, excelling in well-defined domains but struggling with diverse expertise requirements. Multi-agent architectures such as AgentVerse \citep{chen2023agentverse} and MetaGPT \citep{hong2023metagpt} distribute responsibilities across specialized entities, enabling parallel processing and domain-specific reasoning for more complex scenarios.

CrewAI \citep{duan2024exploration} represents a significant advancement in multi-agent implementation, offering organized role-based collaboration through structured task allocation and defined communication protocols. This framework enables agents to perform specialized functions while maintaining coherent system behavior through established workflows. Similarly, LangGraph \citep{duan2024exploration} enhances agent interaction through graph-based state representation, providing persistent memory and conditional branching capabilities that traditional sequential frameworks lack. {Other frameworks like AgentNet \citep{yang2025agentnet} explore decentralized coordination and integrate RAG to allow agents to evolve and specialize, highlighting a trend towards more dynamic and scalable agent systems.}

Despite these frameworks' significant advancements in multi-agent collaboration and task decomposition, application gaps remain in assistive technology for the visually impaired. {While emerging research explores using LLM agents as context-aware UI adjusters for visually impaired users \citep{tang2025exploring},} existing systems primarily focus on general task processing or commercial applications, lacking specialized designs for the unique needs of visually impaired groups. Particularly in scenarios requiring comprehensive environmental understanding and real-time response, existing frameworks often lack targeted optimization and adaptation.

This study applies multi-agent frameworks to create assistive technology for visually impaired users, implementing specialized agents that handle visual perception, language processing, and contextual reasoning within a unified system. Through flow-based workflow management, the system achieves scene type-based task allocation while maintaining coherent user experiences, providing visually impaired users with more natural and fluid environmental information acquisition methods, filling the application gap of multi-agent systems in visually impaired assistive technology.

\subsection{Retrieval-Augmented Generation} 
Retrieval-Augmented Generation(RAG) \citep{lewis2020retrieval} is a method that combines retrieval systems with generative models by retrieving relevant information from external knowledge bases to enhance language model outputs \citep{xiong2024benchmarking}. The RAG architecture primarily consists of four components: knowledge base indexing, query processing, retrieval engine, and generation model, addressing the limitations of large language models' knowledge and factual errors.

study in medical and legal domains has demonstrated the effectiveness of RAG. Medical studies show that RAG can improve large language model accuracy by up to 18\% \citep{xiong2024benchmarking}, effectively integrating professional knowledge \citep{alonso2024medexpqa}. In legal consultation scenarios, multi-turn dialogue RAG systems can precisely retrieve relevant legal documents and generate professional responses \citep{li2025lexrag}.

{Although traditional RAG primarily applies to text, recent work has begun extending its core principles to multimodal tasks. For example, V-RAG \citep{chen2025document} was proposed as a vision-centric RAG framework to reason over thousands of visual documents, and RAVQA-VLM \citep{bhat2025expertneurons} integrates RAG with a VLM to answer questions about scientific figures by retrieving context from source papers. This trend of applying RAG to vision-language contexts validates the approach of using retrieval to enhance multimodal reasoning.} This research applies the RAG concept to visual scene assistance, constructing a vectorized memory system that enhances current scene understanding through historical scene text description retrieval. This approach effectively compensates for the limitations of single-perspective perception, providing visually impaired users with more comprehensive environmental information. Through the integration of historical scene knowledge, the system can provide spatial information beyond the current field of view, helping visually impaired users build more complete environmental mental models.

\section{Methods}\label{Sec_3}
\subsection{Preliminary}
A comprehensive solution is proposed that includes both a vision-language model quantization strategy based on GPTQ method and a multi-agent deployment framework specifically designed to assist visually impaired individuals. The approach effectively compresses VLMs for deployment in computationally constrained environments while maintaining their ability to perform complex assistive tasks. VLMs typically consist of a visual encoder $\phi_V$, a language encoder $\phi_L$, and a multimodal fusion module $\phi_M$. The visual encoder maps an input image $\mathbf{I} \in \mathbb{R}^{H \times W \times 3}$ to visual features $\mathbf{F}_V \in \mathbb{R}^{N_v \times D_v}$, where $N_v$ is the length of the visual feature sequence and $D_v$ is the feature dimension. The language encoder processes text input $\mathbf{T}$ to generate language features $\mathbf{F}_L \in \mathbb{R}^{N_l \times D_l}$, where $N_l$ is the language sequence length and $D_l$ is the feature dimension. The multimodal fusion module integrates these features into a unified representation $\mathbf{F}_M \in \mathbb{R}^{(N_v+N_l) \times D_m}$ for multimodal tasks.

GPTQ quantization is a Hessian-based one-shot quantization technique originally designed for LLMs. For a weight matrix $\mathbf{W} \in \mathbb{R}^{m \times n}$, GPTQ obtains the quantized weights $\mathbf{W}_q$ by solving:
\begin{flalign}\label{A_5}
\mathbf{W}_q = \arg \min_{\mathbf{W}' \in Q} \|\mathbf{W} - \mathbf{W}'\|^2_F
\end{flalign}
To reduce quantization error propagation, GPTQ adopts a column-by-column strategy:
\begin{flalign}\label{A_6}
\min_{\mathbf{q}_j} \|\mathbf{x}_j - \mathbf{W}_j\mathbf{q}_j\|_{\mathbf{H}_j}^2
\end{flalign}
where $\mathbf{q}_j$ represents the quantized value of the $j$-th column of weight matrix $\mathbf{W}$, $\mathbf{x}_j$ is the input feature corresponding to the $j$-th column, $\mathbf{W}_j$ is the original $j$-th column of the weight matrix, and $\mathbf{H}_j$ denotes the Hessian matrix that weighs the quantization error to minimize performance degradation.

{Despite its success in language models, directly applying GPTQ to multimodal large models presents unique challenges. While recent methods have begun to address data-level discrepancies between modalities, optimizations for the fundamental sensitivity differences between the architectural components themselves remain underexplored.}

The multi-agent framework leverages the quantized vision-language model to address specific needs of visually impaired users. The system comprises specialized agents that work collaboratively to process visual information, understand user queries, and provide contextually relevant assistance. These agents interact through a coordinated message passing system, represented as a directed graph $G = (V, E)$ optimized for assistive tasks such as scene description, navigation guidance, and text recognition in natural environments.

RAG is a method that combines retrieval systems with generative models by retrieving relevant information from external knowledge bases to enhance language model outputs. In this research, the core principles of RAG can be formally represented as:
\begin{flalign}
P(y|x) = \sum_{z \in Z} P(y|x,z)P(z|x)
\end{flalign}
where $x$ is the user query, $y$ is the generated answer, and $z$ is the relevant document retrieved from knowledge base $Z$. This formula indicates that the final answer probability distribution is a weighted sum of the answer probability under each retrieved document $z$ and the relevance of that document.

The retrieval process in RAG systems can be implemented through vector similarity calculations:
\begin{flalign}
sim(q, d_i) = \frac{E(q) \cdot E(d_i)}{||E(q)|| \cdot ||E(d_i)||}
\end{flalign}
where $E(q)$ is the vector representation of the query, and $E(d_i)$ is the vector representation of document $d_i$. The system selects the $k$ documents with the highest similarity as context:
\begin{equation}
Z_k = \text{top-k}_{d_i \in D}(sim(q, d_i))
\end{equation}

{The remainder of this section details the two technical components designed to achieve this goal. First, we will describe the model quantization methods that enable efficient deployment. Second, we will detail the multi-agent framework that provides the assistive logic. We will emphasize how these components interoperate to create an effective assistance system, with a specific focus on how RAG techniques are leveraged to enhance environmental understanding and knowledge acquisition.}

\subsection{Design Cross-Modal Differentiated Quantization Framework}
{To efficiently quantize large VLMs, this work introduces the CMDQ framework, which combines modality-specific processing strategies with optimized computational kernels. This differentiated approach proved critical for preserving model performance in our experiments, outperforming naive GPTQ methods.}This design implements differentiated quantization for the distinct characteristics of visual encoders and cross-modal processing components, while maintaining computational performance through efficient Triton implementation and compact storage formats, and providing performance analysis support via NVIDIA NVTX (NVIDIA Tools Extension) technology, effectively balancing model accuracy and resource utilization, offering a reliable technical foundation for the deployment of large-scale multimodal models.

\subsubsection{Design Modality-Specific Module Partitioning Strategy}
To address the special structure of VLMs, a modular quantization strategy was designed, dividing the model into vision encoding modules and cross-modal processing modules for differentiated quantization processing. CogVLM2 \cite{hong2024cogvlm2} was used as the experimental baseline model to verify the effectiveness of this strategy on large-scale VLMs.

In practice, calibration data for vision and multimodal modules is collected through custom VisionCatcher and MultimodalCatcher modules, ensuring that the quantization process is based on the true distribution characteristics of each modality. The vision calibration dataset can be represented as:
\begin{flalign}
    \mathbf{D}^V = \{\mathbf{x}_1^V, \mathbf{x}_2^V, ..., \mathbf{x}_N^V\}
\end{flalign}
where $\mathbf{x}_i^V \in \mathbb{R}^{S_V \times D_V}$ represents the vision features of the $i$-th sample, $S_V$ is the vision sequence length $(image\_size / pat$ $ch\_size)^2 + 1$, $D_V$ is the vision embedding dimension, and $N$ is the number of calibration samples. Similarly, the multimodal calibration dataset can be represented as:
\begin{flalign}
    \mathbf{D}^M = \{(\mathbf{x}_1^M, \mathbf{M}_1, \mathbf{P}_1, \mathbf{T}_1), ..., (\mathbf{x}_N^M, \mathbf{M}_N, \mathbf{P}_N, \mathbf{T}_N)\}
\end{flalign}
where $\mathbf{x}_i^M \in \mathbb{R}^{S_M \times D_M}$ represents the multimodal features, $\mathbf{M}_i$, $\mathbf{P}_i$, and $\mathbf{T}_i$ are the attention mask, position embedding, and token type embedding, respectively.

The quantization process adopts a clear modular separation strategy: first complete the quantization of the vision encoder layers, then process the cross-modal processing modules. Each module uses its own independent calibration data for quantization, avoiding potential interference from cross-modal quantization. For the weight matrix $\mathbf{W}_i^V$ in the vision encoder layer $i$, the quantization process can be expressed as:
\begin{equation}
\begin{split}
    Q(\mathbf{W}_i^V) = s_i^V \cdot \text{round}\left(\frac{\mathbf{W}_i^V}{s_i^V}\right) + z_i^V
\end{split}
\end{equation}
where $s_i^V$ is the scaling factor, and $z_i^V$ is the zero-point offset, which are calculated through vision calibration data.

For the cross-modal module of CogVLM2, a component grouping strategy for quantization is adopted. Specifically, the components of the cross-modal layer are divided into four processing groups: (1) attention query-key-value projection group, including modality-specific expert components; (2) attention output projection group; (3) MLP gate and upper projection group; (4) MLP lower projection group. For the weight matrix $\mathbf{W}_{j,g}^M$ in the $j$-th cross-modal layer's $g$-th group, the quantization process is:

 \begin{flalign}
    Q(\mathbf{W}_{j,g}^M) = s_{j,g}^M \cdot \text{round}\left(\frac{\mathbf{W}_{j,g}^M}{s_{j,g}^M}\right) + z_{j,g}^M
\end{flalign}  
Each group of components shares the same quantization process; for different components $c$ in the same group $g$, the scaling factors satisfy:
\begin{flalign}
s_{j,g,c_1}^M \approx s_{j,g,c_2}^M \approx ... \approx s_{j,g,c_n}^M
\end{flalign}

This ensures that function-related components receive consistent quantization processing. It is worth noting that although these groups are processed in order, each group uses the same original calibration input data, not the output of the previous group's quantization. There is also no direct relationship between the quantization parameters of the vision and cross-modal modules:
\begin{flalign}
\{s_i^V, z_i^V\} \perp \{s_{j,g}^M, z_{j,g}^M\}
\end{flalign}

This modular quantization strategy {ensures} the quantization process{es} of different functional modules {are} independent, {mitigating the risks associated with applying a single, uniform} strategy to the entire model {and thereby} maintaining {performanc}e on different modality data. {While} the {current} implementation is based on the CogVLM2 model, the core principle of this method {is generalizable} to other VLMs with similar multimodal architectures.

\begin{figure*}[htbp]
    \centering
    \includegraphics[width=0.9\textwidth]{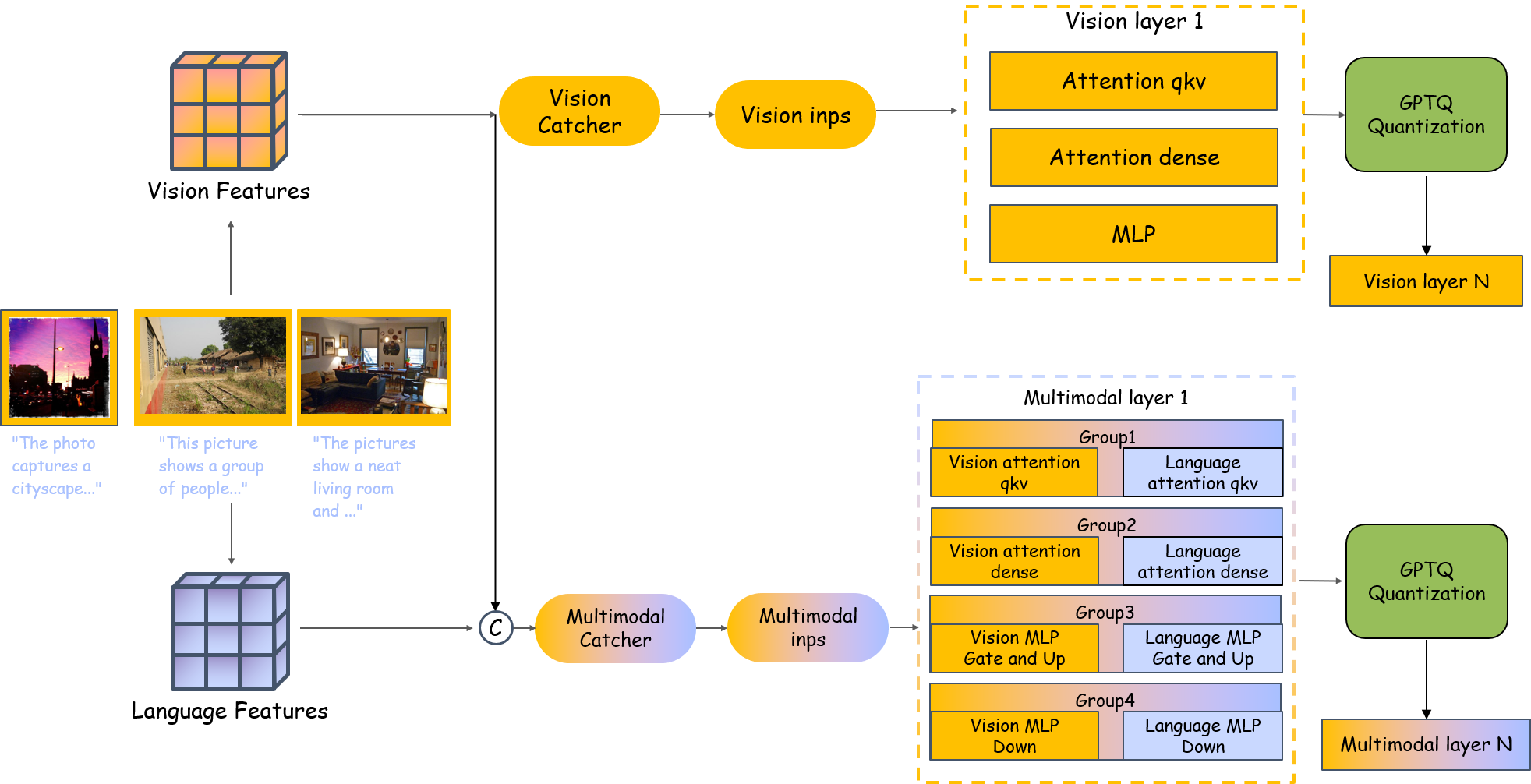} 
\caption{Illustration of the modality-specific module partitioning strategy for quantization in VLMs. The figure depicts the process of collecting calibration data using VisionCatcher and MultimodalCatcher modules, followed by the quantization of vision encoder layers and cross-modal processing modules separately. The modular quantization approach ensures that each functional module undergoes independent quantization, reducing interference and maintaining model performance across different modalities.}
 \label{fig:overall}
\end{figure*}

\subsubsection{Design Efficient Dequantization Computation and Storage Optimization}
To improve the inference efficiency of quantized models, a comprehensive approach was developed that addresses both computation and storage aspects. The solution combines an efficient Triton-based dequantization kernel with a compact storage format, {resulting in high inference speed while minimizing storage requirements. This optimization was essential in reducing the 19B model's footprint from approximately 38GB to 11.3GB.}

The bit-packed storage format optimizes model storage and memory bandwidth usage by tightly packing $N$-bit quantized weight values into 32-bit integers:
\begin{flalign}
f_{int} = \frac{32}{N}
\end{flalign}

This approach compresses the model size to $N/16$ times that of the original FP16 model, achieving $4\times$ compression with 4-bit quantization. The packing process uses bit shifting operations:
\begin{flalign}
\mathbf{qw}[row] = \mathbf{iw}[j] \ll (N \times (j - i))
\end{flalign}
where $\mathbf{qw}$ is the compressed weight tensor, $\mathbf{iw}$ is the quantized integer weight vector, and $\ll$ represents the bitwise left shift operation. The bitwise left shift operation moves all bits in the binary representation of a number to the left by a specified number of positions, effectively multiplying the number by a power of 2 for each position shifted.

To support the group-wise quantization strategy described earlier, an efficient parameter storage structure was implemented for each quantization group:
\begin{equation}
\begin{split}
 \text{scales} &\in \mathbb{R}^{G \times O}\\
\text{qzeros}  &\in \mathbb{Z}^{G \times (O/32 \times N)}\\
\text{g\_idx}  &\in \mathbb{Z}^{I}
\end{split}
\end{equation}
where $G = \lceil I / \text{groupsize} \rceil$ is the number of groups, $I$ is the input feature dimension, $O$ is the output feature dimension, and $N$ is the number of quantization bits. This design makes the storage overhead proportional to the number of groups rather than weights, substantially reducing memory requirements.

Zero-point offsets were further optimized using bit-packed storage:
\begin{equation}
    \begin{split}
        \text{qzeros}[:, col] =& \text{qzeros}[:, col] | (\text{zeros}[:, j] \\
        &\ll (N \times (j - i)))
    \end{split}
\end{equation}
where $\ll$ represents the same bitwise left shift operation described earlier, which shifts bits to the left by the specified number of positions.

This reduces zero-point storage to $N/32$ times the original requirements.

The dequantization algorithm efficiently unpacks quantized values through bitwise operations:
\begin{equation}
    \begin{split}
        sh &= (i \bmod f_{int}) \times N \\
        value &= (\text{pack\_value} \gg sh) \& (2^N - 1) \\
        dq &= (value - zero) \times scale 
    \end{split}
\end{equation}
where $\gg$ represents the bitwise right shift operation and $\&$ represents bitwise AND, $sh$ is the shifter value. The bitwise right shift operation moves all bits in the binary representation of a number to the right by a specified number of positions, effectively dividing the number by a power of 2 for each position shifted.

To maximize GPU utilization, multi-level parallelism with automatic tuning was implemented:
\begin{equation}
    \begin{split}
        \text{block\_config} &= \{M, D, K\}\\
        \text{thread\_config} &= \{W_n, S_n\}
    \end{split}
\end{equation}
where $\text{block\_config}$ defines matrix tiling dimensions, with $M$ representing output rows, $D$ representing output columns, and $K$ representing the inner product dimension, while $\text{thread\_config}$ specifies warp counts $W_n$ and pipeline stages $S_n$.

Optimal configuration is selected pre-compilation:
\begin{equation}
    \begin{split}
        \text{configs} &= \{\text{config}_1, \text{config}_2, \ldots, \text{config}_n\}\\
        \text{optimal\_config} &= \underset{c \in \text{configs}}{\text{argmin}}\ \text{execution\_time}(c, M, D, K)
    \end{split}
\end{equation}
{By evaluating} these configurations through pre-compilation, {the system can select} the computational strategy that best fits the current hardware and model structure. {The synergistic interplay of} storage optimization and dequantization computation optimization {provides} a complete efficient inference solution for large-scale VLMs.

The block computing approach partitions the matrix multiplication operation into manageable tiles to optimize cache utilization and minimize memory access latency. For an input matrix $A \in \mathbb{R}^{M \times K}$ and a quantized weight matrix $B \in \mathbb{R}^{K \times D}$, the computation is divided into blocks of size $B_M \times B_D \times B_K$. 

When executing quantized matrix multiplication, the entire computation space is partitioned into a grid of thread blocks, with each thread block responsible for computing a specific sub-block $C_{sub} \in \mathbb{R}^{B_M \times B_D}$ of the output matrix $C \in \mathbb{R}^{M \times D}$. 

The thread block loads a tile of $A$ of size $B_M \times B_K$ into shared memory, while simultaneously loading and performing dequantization on a tile of $B$ of size $B_K \times B_D$ using the previously described quantization parameters, then performs the local matrix multiplication $C_{ij} += \sum_{k=1}^{B_K} A_{ik} \cdot B_{kj}$. After computation, the results are accumulated and stored back to the corresponding location in global memory. 

Through parallel processing of all thread blocks, the complete output matrix $C$ is computed, achieving efficient distributed computation. This approach fully utilizes the GPU's parallel computing capabilities, ensuring high arithmetic intensity and efficient memory bandwidth utilization, achieving near-optimal performance across various matrix dimensions.

\begin{figure*}[htbp]
    \centering
    \includegraphics[width=0.9\textwidth]{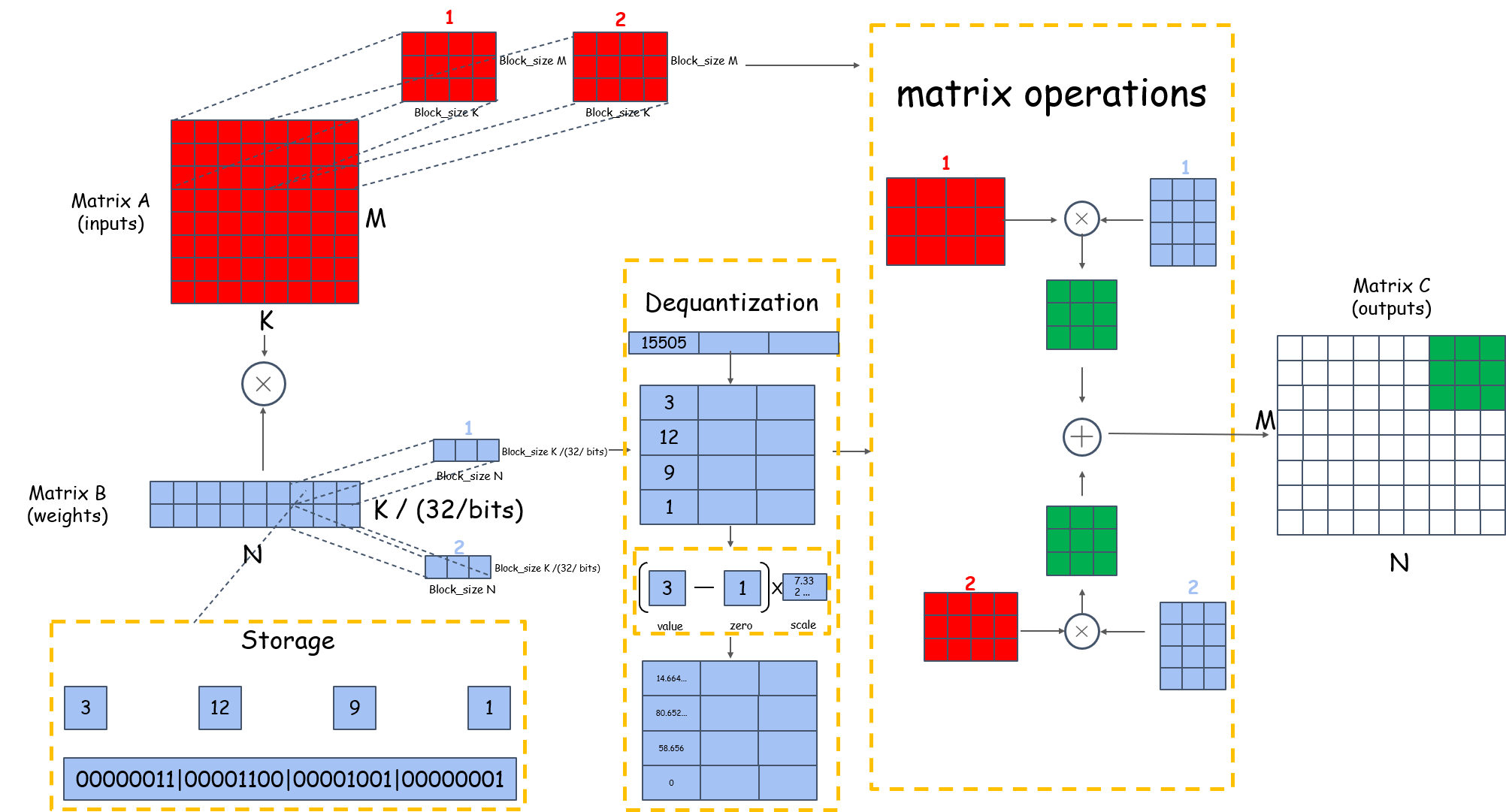} 
\caption{Efficient dequantization computation and storage optimization architecture. The diagram illustrates the bit-packed storage format (bottom left) with multiple weights packed into 32-bit integers, the dequantization process (center) with unpacking, scaling, and zero-point adjustment operations, and the matrix multiplication workflow (right) that enables efficient computation. This optimized architecture reduces memory requirements while maintaining computational efficiency for large vision-language models.}
 \label{fig:overall}
\end{figure*}

\subsubsection{Performance Analysis and Optimization Framework Design}
{We constructed a fine-grained analysis and optimization framework, based on NVIDIA NVTX technology, to evaluate quantization efficiency and guide optimization. This framework enabled detailed performance profiling by implementing a hierarchical tracking system that marks key computational stages, such as the dequantization computation and bias addition within the quantized linear layers. This granular analysis provided intuitive performance insights and precise optimization guidance. Leveraging this NVTX data, we identified key bottlenecks and subsequently developed specialized quantization processes for our multimodal model. This analysis framework was instrumental in understanding model performance and providing reliable technical support for the efficient deployment of large-scale, complex visual-language models.}

To facilitate understanding of the comparative advantages of the CMDQ, Table \ref{tab:quant_methods} provides a comprehensive comparison of different quantization methods. Existing approaches like ZeroQuant \cite{yao2022zeroquant}, AdaRound \cite{nagel2020up}, SmoothQuant \cite{xiao2023smoothquant}, Q-VLM \cite{wang2024q}, GPTQ \cite{frantar2022gptq}, and Bitsandbytes \cite{dettmers2022gpt3} have made significant contributions to the field of model quantization. However, they exhibit certain limitations when applied to VLMs. For instance, some methods lack support for cross-modal differentiated quantization, while others fail to provide pre-computed quantization capabilities or efficient dequantization processes. The proposed CMDQ method fills these gaps by integrating training-free, fast large-model quantization with cross-modal differentiation, pre-computed quantization, VLM support, a modular quantization architecture, and efficient dequantization. Furthermore, it achieves an effective balance between accuracy and efficiency, making it a more comprehensive solution for large VLM deployment.

\begin{table*}[htbp]
\centering
\caption{Comparison of Different Quantization Methods}
\label{tab:quant_methods}
\begin{tabular}{l|cccccccc}
\toprule
\textbf{Method} & \makecell{Training\\Free} & \makecell{Fast Large\\Model\\Quantization} & \makecell{Cross-Modal\\Differentiate} & \makecell{Pre-computed\\Quantization} & \makecell{VLM\\Support} & \makecell{Modular\\Quantization\\Architecture} & \makecell{Efficient\\Dequantization} & \makecell{Accuracy-\\Efficiency\\Balance} \\
\midrule
ZeroQuant \cite{yao2022zeroquant} & $\checkmark$ & $\times$ & $\times$ & $\checkmark$ & $\times$ & $\times$ & $\checkmark$ & $\checkmark$ \\
AdaRound \cite{nagel2020up} & $\checkmark$ & $\times$ & $\times$ & $\checkmark$ & $\times$ & $\times$ & $\times$ & $\checkmark$ \\
SmoothQuant \cite{xiao2023smoothquant} & $\checkmark$ & $\checkmark$ & $\times$ & $\times$ & $\times$ & $\times$ & $\checkmark$ & $\checkmark$ \\
Q-VLM \cite{wang2024q} & $\checkmark$ & $\checkmark$ & $\checkmark$ & $\checkmark$ & $\checkmark$ & $\times$ & $\times$ & $\times$ \\
GPTQ \cite{frantar2022gptq} & $\checkmark$ & $\checkmark$ & $\checkmark$ & $\checkmark$ & $\times$ & $\times$ & $\checkmark$ & $\checkmark$ \\
Bitsandbyes \cite{dettmers2022gpt3} & $\checkmark$ & $\checkmark$ & $\checkmark$ & $\times$ & $\checkmark$ & $\checkmark$ & $\checkmark$ & $\checkmark$ \\
Ours (CMDQ) & $\checkmark$ & $\checkmark$ & $\checkmark$ & $\checkmark$ & $\checkmark$ & $\checkmark$ & $\checkmark$ & $\checkmark$ \\
\bottomrule
\end{tabular}
\end{table*}

\subsection{Design Multi-Agent Assistance Framework}
{To address the fundamental challenge of limited single-perspective perception for visually impaired individuals, we designed a flow-based multi-agent assistance framework engineered to provide holistic environmental understanding that transcends the immediate field of view. This framework, built on CrewAI's Flow execution model and powered by the quantized VLM, establishes a robust system for real-time capture, analysis, and interpretation of visual scenes. By integrating three core components—dynamic scene classification, a vectorized memory system, and multimodal interaction—the framework intelligently processes diverse scenarios, including text recognition, obstacle detection, and environmental description. The system orchestrates agent collaboration through router and listener patterns. Critically, it implements persistent storage and efficient retrieval of scene memories via ChromaDB, enabling the system to leverage historical knowledge. This vectorized memory is the key to achieving cross-perspective understanding, allowing the system to provide context and information beyond what is currently visible. Finally, the framework's built-in streaming speech synthesis and recognition mechanisms ensure a natural and low-latency interaction experience. This multi-layered agent collaboration architecture not only improves the accuracy of scene understanding but also significantly reduces system response latency through the strategic reuse of historical data, providing timely and precise environmental perception assistance that overcomes the limitations of sequential, single-frame analysis.}
\subsubsection{Design Flow-Based Agent System Architecture}
The multi-agent assistance framework designed in this study is built on the CrewAI framework, with VLM serving as its foundation, adopting a declarative flow design pattern that orchestrates and controls agent behaviors through flow control and event-driven mechanisms. The entire system can be formally represented as a directed graph $G=(V,E)$, where $G$ represents the entire flow graph, $V$ is the set of nodes (functional modules), and $E$ is the set of directed edges (control flow).

The system's execution process is divided into five main stages: image capture and basic analysis, scene change detection, scene type classification, specific scene processing, and user interaction dialogue. This design forms a complete perception-analysis-interaction closed loop, enabling the system to continuously provide environmental understanding support for visually impaired users. In the image capture stage, the system continuously acquires a sequence of three images $I = \{I_1, I_2, I_3\}$, and extracts scene descriptions, object lists, and behavior information from each frame, constructing multimodal feature representations. The scene change detection stage determines whether significant changes have occurred in the environment by comparing and analyzing these three frames, which can be represented as the function $\text{SceneChange}(I_1, I_2, I_3) \rightarrow {true,false}$. When scene changes are detected, the system enters the scene classification stage, categorizing the current environment into a specific type; if no changes are detected, it continues environmental monitoring, forming a closed-loop control.

The scene classification system implements a multimodal analysis approach that maps visual inputs to three specialized processing pathways: text recognition scenes, obstacle detection scenes, and environmental description scenes. The classification decision can be represented as $\text{SceneClassify}(I_1, I_2, I_3, O_3, A_3) \rightarrow {T, O, D}$, where $I_1, I_2, I_3$ are the descriptions of three image frames, $O_3$ is the list of objects in the current image, $A_3$ is the list of actions in the current image, and the output values represent text scenes $(T)$, obstacle scenes $(O)$, and description scenes $(D)$ respectively. The system guides VLMs to make scene type determinations through carefully designed prompt templates, considering factors such as text presence, navigation hazards, and environmental complexity.

{Following the scene-type determination, the system initiates a specialized processing phase, where tailored analysis is performed for each scene category. In the case of text-focused scenes, the system detects and interprets textual components, delivering details on their content, positioning, and relevance. For obstacle-dense scenes, the system pinpoints potential navigational risks and viable routes, subsequently offering guidance to the user. Conversely, for descriptive scenes, the system examines the spatial arrangement, prominent landmarks, and ambient characteristics to assist users in forming a cognitive map of the environment. The outputs from these varied analyses are saved in a structured format, establishing a basis for all future interactions and dialogue.}

After completing scene analysis, the system enters the user interaction dialogue stage, transforming technical analysis results into natural language responses. Based on specially designed conversation templates, the system generates dialogue responses for different scene types, providing information and guidance that meet the needs of visually impaired users. This process can be represented as $R = \mathcal{T}(A, H, L, C_r)$, where $R$ is the generated response, $\mathcal{T}$ is the template function, $A$ represents the analysis results, $H$ is the conversation history, $L$ is the target language parameter, and $C_r$ is the relevant historical scene analysis results retrieved from the vector database {. This RAG mechanism is the core component enabling the system's cross-perspective understanding, which was validated in our retrieval experiments.}. This design enables the system to integrate historical scene knowledge on the basis of current scene understanding, thereby providing more comprehensive and accurate environmental information. At the same time, the system vectorizes the current scene's analysis results and stores them in ChromaDB, forming a continuously expanding scene memory library for future scene understanding reference (this vectorized memory system and scene retrieval mechanism will be discussed in detail in the next section).

The system's decision process is implemented based on router and listener patterns, forming a flexible state transition network. The listener pattern defines sequential dependencies between functional units, ensuring that the system executes processes in a predetermined order, which can be represented as $f_j \circ f_i(s) = f_j(f_i(s))$, where $f_i$ must complete before $f_j$ begins processing. The router pattern implements conditional branching logic, allowing the system to choose different execution paths based on analysis results, represented as $r(s) = \begin{cases} p_1 & \text{if } \gamma(s) = \text{true} \\ p_2 & \text{if } \gamma(s) = \text{false} \end{cases}$, where $r$ is the routing function, $s$ is the current state, $\gamma$ is the evaluation function, and $p_1, p_2$ are alternative processing paths. This mechanism implements flexible conditional control flow, enabling agents to make appropriate decisions based on environmental states.

Throughout the entire processing workflow, the system maintains contextual awareness through state management mechanisms, including cross-scene analysis results and conversation history. While existing multi-agent frameworks such as AgentVerse and MetaGPT have established important foundations for agent collaboration, they typically lack specialized workflows for visual assistance scenarios. The flow-based architecture extends these approaches through scene-specific processing paths and declarative control flows specifically designed for vision-based assistance tasks. By implementing router and listener patterns tailored to environmental perception needs, this design allows the system to handle complex environments in continuous monitoring-analysis-response cycles, providing more responsive and adaptive assistance for visually impaired users than previously possible with general-purpose agent architectures. The system's cyclic execution process embodies the continuity of environmental monitoring, enabling agents to adapt to environmental changes and provide corresponding support.

\begin{figure*}[htbp]
    \centering
    \includegraphics[width=0.9\textwidth]{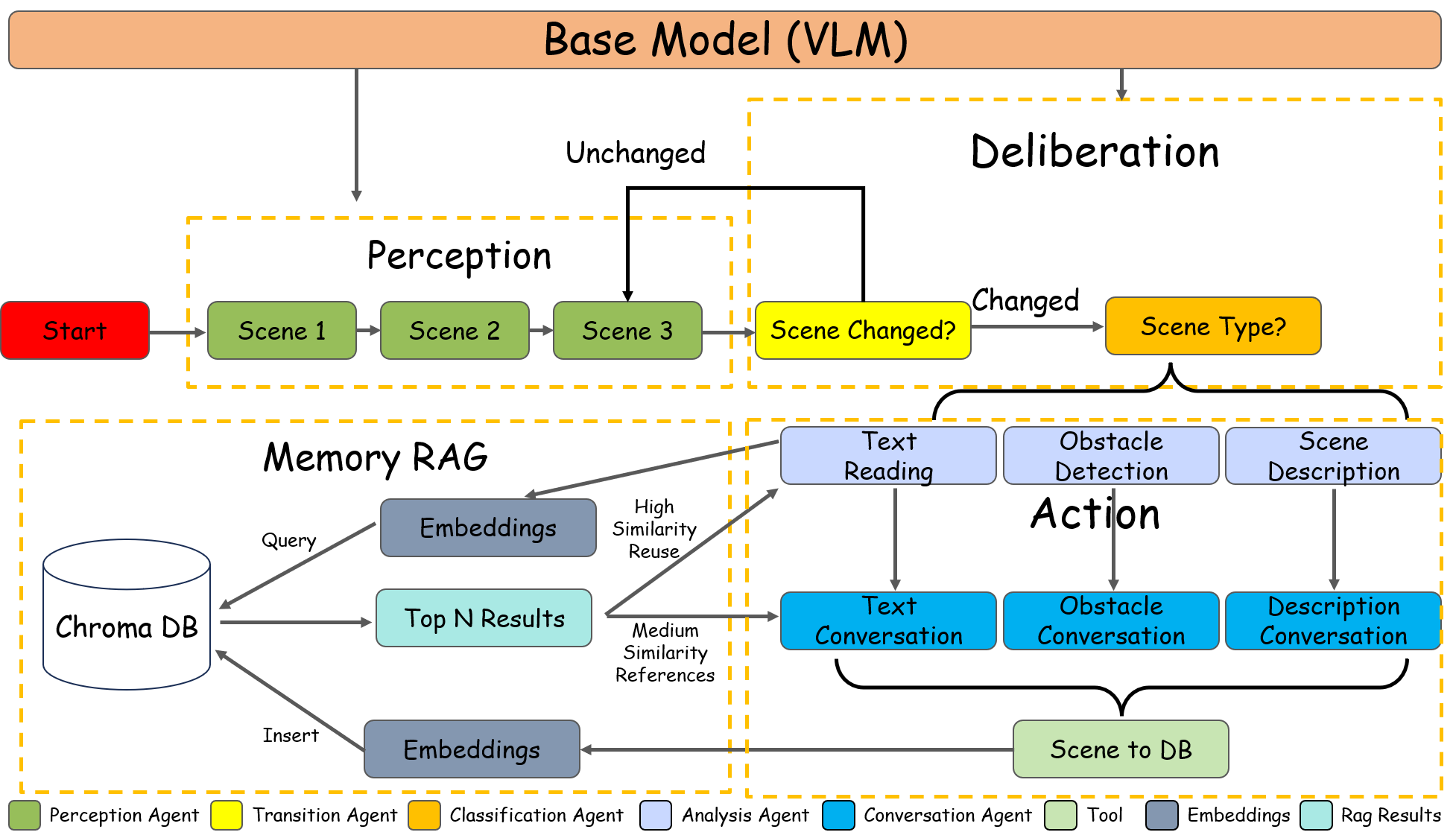} 
\caption{Flow-based Multi-agent Visual Assistance Framework. The architecture integrates Perception (scene capture), Memory RAG (similarity-based scene retrieval), and Deliberation (analysis and interaction) modules, all powered by VLM. The system optimizes processing by directly reusing analysis for high-similarity scenes and leveraging historical data for medium-similarity scenes, creating an efficient closed-loop system for visually impaired users.}
 \label{fig:overall}
\end{figure*}

\subsubsection{Design Vectorized Memory System and Scene Retrieval Scheme}
{The RAG-based vectorized memory system in this research is designed to overcome two primary obstacles: first, the inherent limitation of a single viewpoint restricts visually impaired users from grasping a comprehensive understanding of their surroundings; second, the significant computational cost of performing redundant scene analyses for every frame negatively impacts the system's responsiveness. By leveraging historical scene data retrieval to augment the current perceptual analysis, the system concurrently expands the user's perceptual scope and boosts operational efficiency.}

The application of the RAG framework in this system includes three key components: vectorized representation, similar scene retrieval, and knowledge fusion. First, the system implements vectorized storage and retrieval of scene memories using ChromaDB. For scene feature vectorization, the system integrates scene descriptions, recognized objects, and detected actions to generate multi-dimensional semantic representations. The feature vector $E(S)$ for a given scene $S$ can be represented as:
\begin{flalign}
E(S)=f_{embed}(D_S \oplus O_S \oplus A_S)
\end{flalign}
where $D_S$ is the scene description text, $O_S$ is the object list, $A_S$ is the action description, $\oplus$ represents text concatenation operation, and $f_{embed}$ is the text embedding model. This representation method effectively captures the core features of scenes, allowing similar scenes to naturally cluster in vector space.
Based on vector representation, the system implements a similar scene retrieval mechanism. Given the current scene $S_{curr}$ and the historical scene collection ${S_1, S_2, ..., S_n}$, the system calculates the L2 distance between them and converts it to similarity:
\begin{equation}
    \begin{split}
        d(S_{curr}, S_i) &= ||E(S_{curr}) - E(S_i)||_2\\
        sim(S_{curr}, S_i) &= 1 - \frac{d(S_{curr}, S_i)}{d_{max}}
    \end{split}
\end{equation}
where $d_{max}$ represents the maximum possible distance between any two scene vectors in the feature space, serving as a normalization factor to scale similarity scores to the [0,1] range.

The system adopts a {dual-filter strategy, first matching the scene type} ($T(S_{curr}) = T(S_i)$) {and then applying a} two-level {similarity} threshold {(high and low) to the results.} {This combined approach was shown in our ablation studies to be essential for filtering irrelevant, low-similarity memory noise and reducing false positives.} This approach divides retrieved {candidates} into two categories: 1. High-similarity scenes ($M_{exact}$): Scenes with similarity scores exceeding the high threshold and matching the current scene type. 2. Medium-similarity scenes ($M_{ref}$): Scenes with similarity scores between the minimum threshold and the high threshold, also matching the current scene type. Based on the retrieval results, the system employs a hierarchical information integration strategy to enhance scene understanding. For high-similarity scenes, the system directly reuses their complete analysis results:
\begin{flalign}
R(S_{curr}) = A(S_{i}), \text{where } S_i \in M_{exact}
\end{flalign}

For medium-similarity scenes, the system uses their analysis results as supplementary context, guiding VLMs to generate enhanced understanding through prompt engineering:
\begin{flalign}
R(S_{curr}) = \text{VLM}(A(S_{curr}), \{A(S_i) | S_i \in M_{ref}\})
\end{flalign}

When both high-similarity and medium-similarity matches exist simultaneously, the system combines the two types of information in the dialogue generation phase, using high-similarity match results as the main body, supplemented by medium-similarity references to provide broader environmental information. This process is not a simple function mapping, but guides VLMs to integrate multi-source information through carefully designed prompt templates, generating coherent and comprehensive environmental descriptions.

This RAG-based scene understanding mechanism is particularly applicable to perspective-limited situations. In indoor environments, as users move, the system can gradually accumulate scene information from different perspectives. When users re-enter similar scenes, the system can not only identify elements in the current field of view but also provide information about objects or spatial layouts that may exist around the current perspective, thereby helping users establish more complete spatial cognition.

{The scene memory repository employs a persistent storage design, cataloging complete metadata for each entry, including scene descriptions, object lists, action information, analysis results, and timestamps. While prior visual assistance systems have been primarily constrained by single-perspective scene understanding, limiting their environmental awareness to the current field of view , our vectorized memory system implements similarity-based retrieval and knowledge fusion mechanisms that explicitly enable cross-perspective reasoning. As users navigate an environment and the system is continuously used, the memory repository dynamically expands, gradually constructing a cognitive mapping of the user's activity space. This approach overcomes the fundamental limitations of traditional perception systems, an advantage that is particularly pronounced in relatively fixed scenes such as homes, offices, or public buildings, by building a repository of accumulated environmental knowledge that improves naturally with system usage. Consequently, the system provides increasingly accurate and comprehensive environmental understanding support over time, without requiring any additional model training or fine-tuning processes.}

\subsubsection{Design Multimodal Interaction and Speech Streaming Processing}
To address the real-time interaction needs of visually impaired individuals during environmental perception, this study designed an efficient multimodal interaction system that transforms complex scene analysis results into natural, fluent speech output and enables two-way interaction through speech recognition. Compared with traditional assistive systems, this system adopts innovative design approaches to reduce interaction latency and optimize user experience.

The system employs a multi-threaded concurrent architecture to implement efficient streaming speech synthesis processing. Unlike existing systems such as VocalEyes \cite{chavan2024vocaleyes} which use lightweight TTS models (such as Parler TTS Mini) to reduce speech synthesis latency but still process complete text segments, the streaming speech approach further reduces time delay. The system implements a streaming speech synthesis processing architecture that decomposes large speech synthesis tasks into smaller subtasks processed in parallel. The entire speech processing system can be formalized as a quaternion $S = (T, P, B, Q)$ where $T$ represents the text processor,$P$represents the audio generator,$B$ represents the buffer, and $Q$represents the task queue collection. The system workflow is based on the producer-consumer model, decomposing large text speech synthesis tasks into small subtasks processed in parallel, effectively reducing end-to-end response latency.

The core of the system consists of two parallel working threads: the TTS conversion thread and the audio playback thread, which communicate through task queues. The TTS conversion thread is responsible for converting text to speech files, representable as the function $f_{tts} : \text{Text} \rightarrow \text{AudioFile}$, while the audio playback thread is responsible for sequentially playing the generated speech files, forming the function $f_{play} : \text{AudioFile} \rightarrow \text{Audio}$.Coordination between the two threads is implemented through an event signaling mechanism:
\begin{equation}
\text{Process}(t) = \begin{cases}
f_{tts}(t) \rightarrow Q_{tts} \\
\text{Wait}(E_{ready}) \circ f_{play}(Q_{play})
\end{cases}
\end{equation}
where $t$ is the input text fragment, $Q_{tts}$ and $Q_{play}$ are the TTS task queue and playback task queue respectively, $E_{ready}$ is the task ready event, and $\circ$ represents the function composition operation. This design enables the system to generate the next audio file in parallel while processing one audio file, significantly reducing the overall response time.

To solve the coordination problem between large language model streaming output and speech synthesis, the system developed a dynamic processing mechanism based on sentence-level text chunking. This mechanism first defines a set of sentence recognition patterns $\mathcal{P} = \{p_1, p_2, ..., p_n\}$, including various sentence boundary patterns such as basic periods, consecutive line breaks, colon endings, etc., implemented through regular expressions. The text buffer $B$ is responsible for accumulating the incoming token stream and identifying complete sentences, and its working process can be formalized as a state transition function:
\begin{equation}
B(s, t) = \begin{cases}
(s + t, \text{null}) & \text{if no pattern matches} \\
(s', \text{sentence}) & \text{if a pattern matches}
\end{cases}
\end{equation}
where $s$ is the current buffer state, $t$ is the newly received text fragment, $s'$ is the text after the sentence boundary, and "sentence" is the identified complete sentence. The first case occurs when no sentence boundary pattern in $\mathcal{P}$  is found in the combined text $(s + t)$, while the second case occurs when a pattern matches, allowing the system to extract a complete sentence and update the buffer. Each time a complete sentence is identified, the system immediately creates a new speech synthesis task using a unique identifier and temporary file path: $\text{Task}(i, t) = (t, f_i, E_i)$, where $i$ is the task identifier,$t$  is the sentence text,$f_{i}$ is the temporary file path, and $E_{i}$ is the ready event signal. This design implements seamless integration of large language model output streams and speech synthesis, allowing users to hear the front part of the speech output before the system completes the full response, effectively reducing user-perceived latency.

{To improve the system's accessibility, this research incorporated several interaction control methods specifically tailored for individuals with visual impairments. An autonomous listening thread constantly monitors for user input, enabling users to halt or manage the audio output through accessible means. Specifically, the system accommodates straightforward button-based controls, like a designated interrupt button to cease the current playback. This permits visually impaired users to manage the flow of information without needing to navigate intricate interface elements. This design accounts for the specific requirements of visually impaired users, offering a more user-friendly and convenient method of interaction.}

The system also adopts optimization measures in temporary resource management, implementing an automatic cleanup mechanism that generates uniquely identified temporary files for each audio task and deletes them immediately after playback, maintaining effective utilization of system resources. Additionally, the system implements flexible speech recognition integration, supporting various speech recognition services, using specially designed error handling and retry mechanisms to improve stability. This mechanism can automatically handle common speech recognition errors, such as environmental noise interference or unclear speech, significantly enhancing system robustness through intelligent retry strategies.

Through these designs, the system provides low-latency, highly available interaction experiences particularly suited to the environmental perception needs of visually impaired users. While existing assistive technologies often struggle with balancing response comprehensiveness and interaction latency, resulting in either incomplete information or prohibitive delays, the approach addresses these fundamental challenges. Previous streaming approaches have typically focused on general text-to-speech applications without considering the unique requirements of visually impaired users in dynamic environments where rapid feedback is essential. The sentence-level chunking mechanism and multi-threaded concurrent processing architecture specifically overcome these limitations, reducing perceived audio response times from over 30 seconds to under 4 seconds while maintaining information quality. This innovation fundamentally transforms the interaction experience for visually impaired users, providing a more natural and fluid way to obtain environmental information through real-time feedback that was previously unachievable with traditional sequential processing methods.

\section{Experiment} \label{Sec_4}
\subsection{Datasets}
Model evaluation was performed exclusively using the VLMEvalKit tool \cite{duan2024vlmevalkit} on two standard multimodal benchmarks: MMBench \cite{liu2024mmbench} and OCR-VQA \cite{mishra2019ocr}, testing the system's performance across different visual understanding scenarios.
MMBench is a systematic benchmark designed to evaluate the capabilities of VLMs. It contains 3,217 multiple-choice questions covering 20 fine-grained ability dimensions, including object recognition, spatial relationships, and complex reasoning tasks. MMBench's hierarchical ability taxonomy categorizes evaluation into perception and reasoning as main categories, further subdivided into more specialized ability categories, providing us with a comprehensive standard for assessing the quantized model's preservation of visual understanding capabilities.
OCR-VQA \ is a dataset focused on testing text recognition and understanding capabilities in images. It contains 207,572 book cover images and over 1 million question-answer pairs distributed across 33 different domains. The diverse text layouts, fonts, and background complexities in book covers provide a rigorous benchmark for evaluating the quantized system's OCR capabilities and text comprehension abilities, enabling us to verify the impact of quantization on text recognition tasks, which is crucial for assessing the practicality of compressed models in real applications.
Together, these datasets provide a reliable testing platform for comprehensively evaluating the impact of cross-Modal differentiated quantization on multimodal large model performance and the practical application effectiveness of the multi-agent system in visual assistance scenarios.

\subsection{Evaluation Metrics}
To comprehensively assess the performance of the cross-modal differentiated quantization framework and multi-agent visual assistance system, specific evaluation metrics tailored to the MMBench and OCR-VQA datasets were employed.

For MMBench, the CircularEval strategy's Top-1 accuracy was adopted as the primary metric. This strategy requires models to correctly answer questions across all choice permutations, providing a more rigorous evaluation than traditional single-pass assessment. Performance was analyzed across six L-2 ability dimensions: Coarse Perception (CP), Fine-grained Perception for single instances (FP-S) and cross instances (FP-C), Attribute Reasoning (AR), Logic Reasoning (LR), and Relation Reasoning (RR). This granular approach enabled precise analysis of how quantization affects different cognitive capabilities.

For the CircularEval strategy, a model is considered successful in solving a question only if it correctly predicts the answer in all circular passes. The metric can be formalized as:

\begin{equation}
    \text{CircularEval}_{\text{acc}} = \frac{\sum_{i=1}^{N} \prod_{j=1}^{c_i} \text{Bool}(p_{i,j} = a_{i,j})}{N}
\end{equation}

\noindent Where $N$ is the total number of questions, $c_i$ is the number of choices for question $i$, $p_{i,j}$ is the model's prediction for question $i$ in circular pass $j$, $a_{i,j}$ is the corresponding ground truth, and $\text{Bool}(p_{i,j} = a_{i,j})$ returns 1 if the condition is true and 0 if false.

For OCR-VQA, overall accuracy and type-specific accuracy metrics were used, focusing on binary questions, book title recognition, author name identification, book genre classification, year recognition, and edition identification. These evaluation dimensions are particularly suitable for validating the capabilities of the text processing agent in the multi-agent framework, which is specifically responsible for recognizing and understanding textual information in scenes, crucial for helping visually impaired individuals comprehend written content in their environment.

\subsection{Experimental Details}
This experiment is based on the CogVLM2-19B multimodal large model, applying the proposed efficient quantization method for model compression. All experiments were conducted on a single NVIDIA RTX 4090 GPU with a memory limit of 24GB. During the quantization process, the original float16 precision linear layers were replaced with the designed quantized linear layers (quantLinear) and the forward propagation function was rewritten using optimized Triton kernels to achieve efficient inference.

{The adopted quantization method utilizes 4-bit (Int4) precision with an asymmetric quantization strategy, configured with a groupsize of 128 and a Hessian damping (percdamp) value of 0.3. For tuning the quantization parameters, a carefully designed calibration dataset is utilized. This dataset is constructed from a fixed subset of 128 samples sourced from the CogVLM-SFT-311K dataset. This sample size was selected as a trade-off between data representativeness and the 24GB hardware memory limit, as preliminary tests indicated that using 256 samples would lead to an out-of-memory (OOM) error. To create this static set, a single image and a corresponding dialogue turn were selected from each source sample during a one-time preprocessing step. The multimodal inputs are subsequently padded to a fixed sequence length of 2048, creating a uniform dataset that ensures experimental reproducibility.
	
The quantization process adopts the cross-modal differentiated strategy. A specially designed data capture mechanism is used to separately obtain the calibration data required for the visual encoder and the cross-modal processing modules. Specifically, this data is used to independently tune the quantization parameters for the visual encoder and the cross-modal processing modules, ensuring the quantized model accurately captures the characteristics of the original weight distribution. This modular approach guarantees the computational independence of the quantization parameters between these distinct architectural components.}

Model evaluation was performed using the VLMEvalKit tool on two standard multimodal benchmarks: MMBench v1.1 TEST and OCRVQA\_TESTCORE. MMBench evaluates visual language models' comprehensive performance across 20 different ability dimensions, while OCRVQA focuses on assessing the model's ability to recognize and understand textual content in images.

The quantization method uses 4-bit (Int4) precision to compress model weights while maintaining high-precision computation for activation values, balancing computational efficiency and model performance. Through this approach, the memory footprint of the 19B parameter model was reduced from approximately 38GB (float16) to only 11.3GB, enabling it to run efficiently on a single consumer-grade GPU. The quantization process adopts a modular design, separately processing the vision encoder and cross-modal processing modules to ensure balanced performance across different modalities. Carefully designed calibration datasets were used to adjust the quantization parameters, ensuring that the quantized model accurately captures the characteristics of the original weight distribution.

\subsection{Experimental Analysis}
As shown in Table \ref{tab:mmbench test}, the Int4 quantized CogVLM2-19B model was compared with several advanced VLMs on the MMBench v1.1 TEST dataset. Evaluation metrics include overall score and performance in various ability dimensions such as Logic Reasoning (LR), Attribute Reasoning (AR), Relation Reasoning (RR), Single-instance Fine-grained Perception (FP-S), Cross-instance Fine-grained Perception (FP-C), and Coarse Perception (CP). The comparison includes models with different parameter scales and memory requirements.

\begin{table*}[htbp]
\centering
\caption{{Performance Comparison of Different VLMs on MMBench v1.1 TEST}}
\label{tab:mmbench test}
\resizebox{0.9\textwidth}{!}{
\begin{threeparttable}
\begin{tabular}{@{}lcc|ccccccc@{}}
\toprule[2pt]
Model & Parameters & Memory & Overall & LR & AR & RR & FP-S & FP-C & CP \\ \midrule
CogVLM2-19B-Chat & 19B & 38GB\tnote{*} & 72.7 & 53.3 & 80.8 & 77.1 & 76.6 & 63.3 & 75.0 \\
\textbf{CogVLM2-19B-Int4 (Ours, CMDQ)} & 19B & 11.3GB & 70.7 & 52.7 & 78.8 & 73.3 & 74.0 & 64.0 & 72.4 \\
CogVLM2-19B-Int4 (Naive GPTQ) & 19B & 11.3GB & 69.5 & 50.2 & 76.3 & 73.0 & 72.9 & 62.7 & 72.1 \\ \hline
XVERSE-V-13B & 13B & 26GB & 72.1 & 42.4 & 73.1 & 72.1 & 79.4 & 67.4 & 78.3 \\
VILA1.5-13B & 13B & 26GB & 71.8 & 47.8 & 79.2 & 69.8 & 77.0 & 65.2 & 76.7 \\
Eagle-X5-13B & 15B & 30GB & 70.0 & 47.8 & 81.6 & 67.1 & 74.0 & 62.2 & 74.4 \\ \hline
Molmo-7B-D & 8B & 16GB & 70.3 & 56.5 & 72.7 & 67.8 & 78.7 & 62.9 & 72.2 \\
Monkey-Chat & 9.8B & 20GB & 69.6 & 47.3 & 82.4 & 58.9 & 75.4 & 63.3 & 75.0 \\
Molmo-7B-O & 8B & 16GB & 69.1 & 50.5 & 72.2 & 70.2 & 73.8 & 63.3 & 72.6 \\
\bottomrule[2pt]
\end{tabular}
\begin{tablenotes}
\small
\item[*] Estimated memory usage of the original FP16 model
\item The quantized model (CogVLM2-19B-Int4) shows only a 2.0\% performance drop compared to the original model while reducing memory requirements by 70.3\%. It achieves comparable performance to models with higher memory footprints and demonstrates advantages over smaller models with comparable memory consumption. All performance data is sourced from the official MMBench \cite{liu2024mmbench} leaderboard.
\end{tablenotes}
\end{threeparttable}
}
\end{table*}

The Int4 quantized model, while maintaining the original 19B parameter scale, significantly reduced memory usage from 38GB to 11.3GB while largely preserving performance similar to the original model (only a 2.05\% decrease). This combination of high compression ratio and minimal performance loss fully demonstrates the effectiveness of the quantization method.

{To further validate the effectiveness of our proposed CMDQ framework, we conducted an ablation study. We removed the CMDQ strategy and instead applied a naive GPTQ method for the same Int4 quantization on the CogVLM2-19B model. As shown in Table \ref{tab:mmbench test}, the model using naive GPTQ achieved an overall score of 69.5 on MMBench, whereas our CMDQ method reached 70.7. This 1.2-point performance improvement, especially the advantages in Attribute Reasoning (AR, 78.8 vs 76.3) and Fine-grained Perception (FP-S, 74.0 vs 72.9; FP-C, 64.0 vs 62.7), clearly demonstrates the significant superiority of the CMDQ framework in handling different modality sensitivities and preserving the model's key capabilities.}

In comparison with other models, the approach shows two significant advantages:

\textbf{Parameter Efficiency Advantage:} Compared to models with 13-15B parameters (such as XVERSE-V-13B, VILA1.5-13B, and Eagle-X5-13B), the quantized model maintains comparable performance while significantly reducing runtime memory requirements. The model performs similarly to Eagle-X5-13B while cutting memory requirements in half. This efficiency is reflected not only in memory usage but also implies lower computational costs and broader deployment possibilities.

\textbf{Performance Advantage:} Compared to models with similar memory usage (16GB), such as Molmo-7B-D and Molmo-7B-O, the quantized model shows comparable or slightly better performance. Despite these models having only 8B parameters and occupying comparable memory space, the model performs similarly overall, demonstrating the efficiency of the method. In Relation Reasoning (RR) ability, the model shows a significant advantage (3.06-14.36 percentage points higher). This indicates that the proposed method can provide stronger model capabilities under the same resource constraints.

In specific ability dimensions, the model performs particularly well in Attribute Reasoning (AR) and Relation Reasoning (RR), achieving high scores of 78.78 and 73.26 respectively, outperforming most comparison models. Even during the quantization process, these complex reasoning abilities were well preserved, proving the robustness of the proposed method.

\begin{table*}[htbp]
\centering
\caption{{Performance Comparison of Different VLMs on OCRVQA\_TESTCORE}}
\label{tab:ocrvqa_performance}
\resizebox{0.9\textwidth}{!}{
\begin{threeparttable}
\begin{tabular}{@{}lcc|cccccc@{}}
\toprule[2pt]
Model & Parameters & Memory & Overall & Cookbooks & Medical Books & History & Reference & Education \\ \midrule
CogVLM2-19B-Chat & 19B & 38GB\tnote{*} & 64.9 & 68.8 & 61.5 & 65.6 & 54.2 & 59.4 \\
\textbf{CogVLM2-19B-Int4 (Ours, CMDQ)} & 19B & 11.3GB & 63.7 & 69.8 & 57.3 & 70.8 & 49.0 & 61.5 \\ 
CogVLM2-19B-Int4 (Naive GPTQ) & 19B & 11.3GB & 62.9 & 68.7 & 57.2 & 67.7 & 48.4 & 59.1 \\
\hline
InternVL-Chat-V1.5 & 26B & 52GB & 64.2 & 68.8 & 60.4 & 63.5 & 46.9 & 53.1 \\
Pixtral-12B & 13B & 26GB & 64.7 & 71.9 & 60.4 & 62.5 & 49.0 & 61.5 \\
ShareGPT4V-13B & 13.4B & 28GB & 64.3 & 68.8 & 61.5 & 69.8 & 43.8 & 56.2 \\ 
\hline
ShareGPT4V-7B & 7.2B & 16GB & 63.4 & 67.7 & 57.3 & 70.8 & 46.9 & 52.1 \\
Mantis-8B-Idefics2 & 8B & 16GB & 62.6 & 71.9 & 54.2 & 42.7 & 62.5 & 55.2 \\
LLaVA-LLaMA-3-8B & 8B & 16GB & 61.3 & 64.6 & 59.4 & 68.8 & 45.8 & 55.2 \\
\bottomrule[2pt]
\end{tabular}
\begin{tablenotes}
\small
\item[*] Estimated memory usage of the original FP16 model
\item The quantized model (CogVLM2-19B-Int4) shows only a 1.2\% performance drop compared to the original model while reducing memory requirements by 70.3\%. It achieves comparable performance to models with higher memory footprints (such as InternVL-Chat-V1.5 and Pixtral-12B) and demonstrates advantages over smaller models with similar memory consumption. All performance data is sourced from the official VLMEvalKit \cite{duan2024vlmevalkit} leaderboard.
\end{tablenotes}
\end{threeparttable}
}
\end{table*}

\begin{figure*}[htbp]
    \centering
    \begin{subfigure}[b]{0.48\textwidth}
        \centering
        \includegraphics[width=\textwidth]{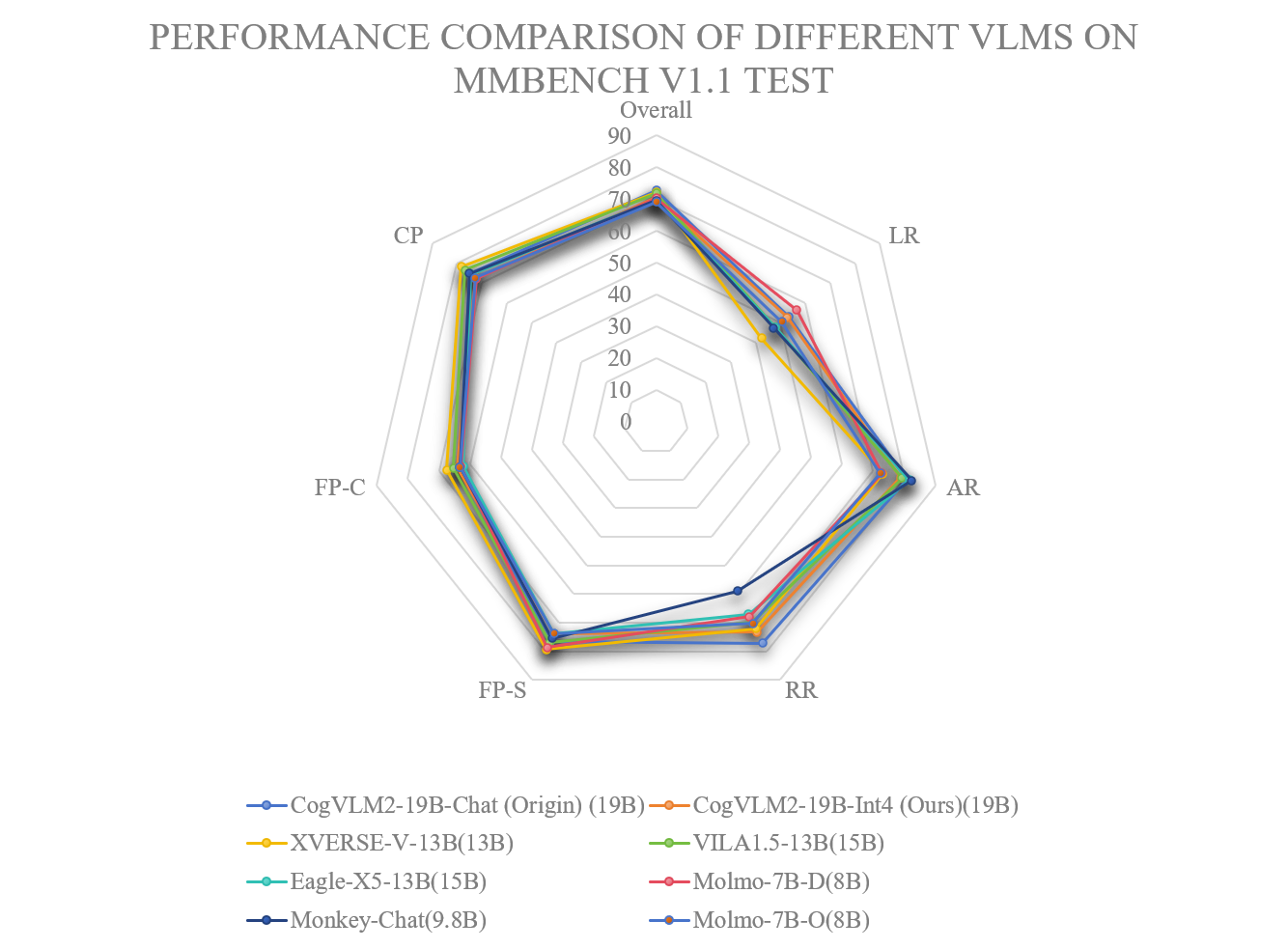}
        \caption{Performance Comparison of Different VLMs on MMBench v1.1 TEST. The Int4 quantized model maintains competitive performance across various ability dimensions while significantly reducing memory requirements.}
        \label{fig:radar_mmbench}
    \end{subfigure}%
    \hfill
    \begin{subfigure}[b]{0.48\textwidth}
        \centering
        \includegraphics[width=\textwidth]{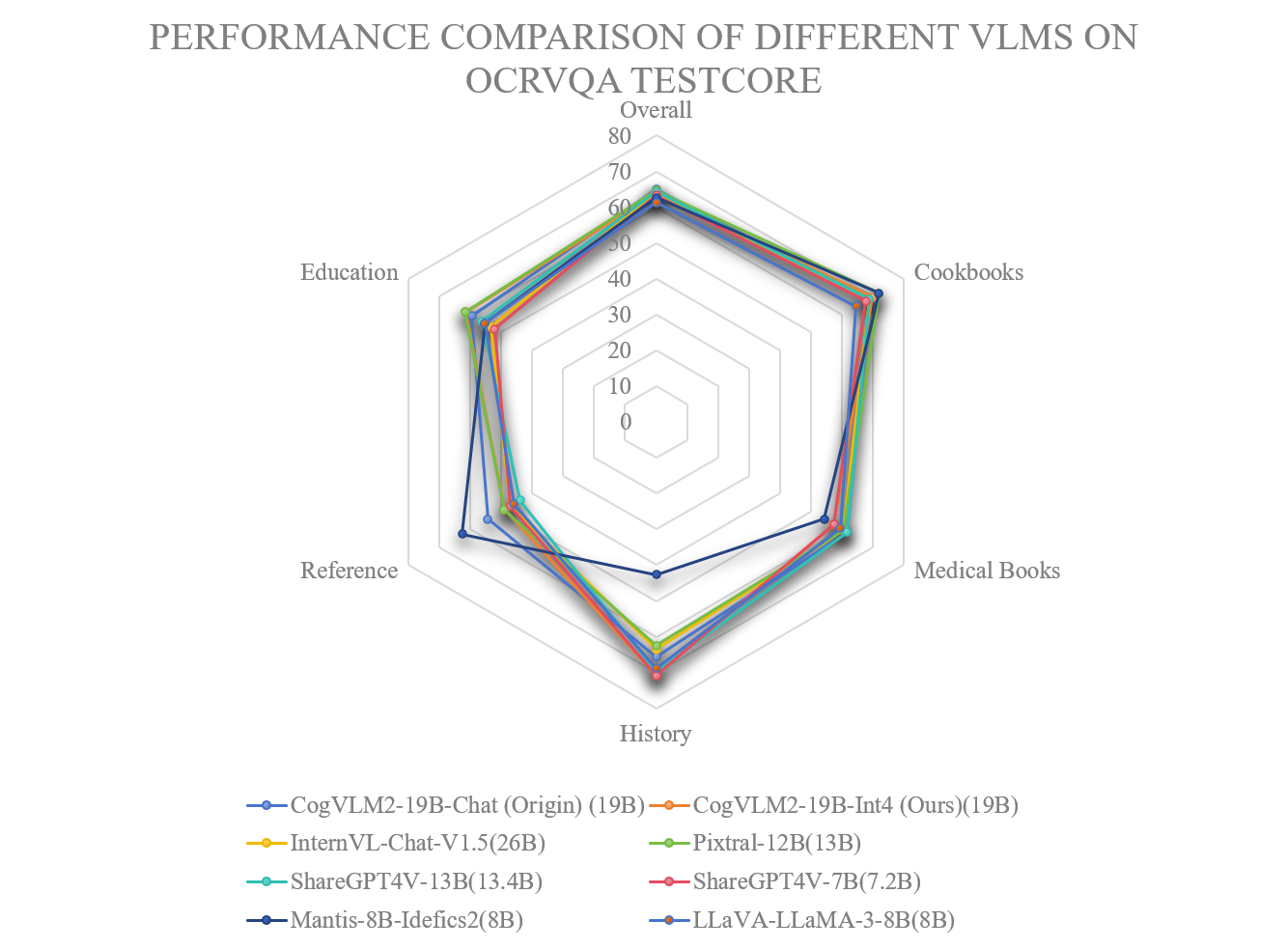}
        \caption{Performance Comparison of Different VLMs on OCRVQA\_TESTCORE. Radar chart illustrates the quantized model's balanced performance across different categories while requiring only $11.3\text{ GB memory}$ compared to larger models.}
        \label{fig:radar_ocrvqa}
    \end{subfigure}
    \caption{Performance comparison of different VLMs on benchmarks. (a) MMBench v1.1 TEST results. (b) OCRVQA\_TESTCORE results.}
    \label{fig:combined_radar}
\end{figure*}

{Table~\ref{tab:ocrvqa_performance} shows the performance of various models on the OCRVQA\_TESTCORE dataset. Our quantized model (Ours, CMDQ) achieved a total score of 63.7 on the OCRVQA test, only 1.2 percentage points lower than the original model (64.9), essentially maintaining similar performance. This was further corroborated by the ablation study: the model quantized with naive GPTQ scored only 62.9, while our CMDQ method was 0.8 points higher. This performance gap was particularly evident in categories such as 'History' (70.8 vs 67.7) and 'Education' (61.5 vs 59.1), again confirming the importance of our differentiated strategy in preserving the model's fine-grained text understanding capabilities. The performance scores reported in this table are collected from the open\_vlm\_leaderboard website \cite{duan2024vlmevalkit}.}

Notably, the quantized model outperforms smaller models with similar memory usage such as ShareGPT4V-7B, while significantly surpassing other small parameter models like Mantis-8B-Idefics2 and LLaVA-LLaMA-3-8B. Compared to the larger (26B) InternVL-Chat-V1.5 model, although the total score is slightly lower, the model reduces memory requirements from 52GB to just 11.3GB, a compression ratio of up to 78.27\%. This significant improvement in resource efficiency enables the model to be deployed on edge devices or portable devices.

In specific category performance, the model shows an uneven distribution of capabilities: scoring 69.79 in Cookbooks, Food \& Wine, 57.29 in Medical Books, 70.83 in History, 48.96 in Reference, and 61.46 in Education \& Teaching. Compared to the original unquantized model, some categories such as History and Education \& Teaching even show improvement, while Medical Books and Reference categories show some decline. {This performance divergence is attributed to the varying sensitivity of tasks to model precision. High-sensitivity categories (Medical/Reference) typically contain high font density, extremely fine-grained text (e.g., complex terminology, subscripts), and complex layouts. Success in these tasks is heavily dependent on the model's ability to capture and reason about these fine visual details. In contrast, low-sensitivity categories (History/Education) rely more on the comprehension of standard text and regular layouts, making them more robust to the information loss introduced by 4-bit quantization. In the current experimental configuration, a uniform 4-bit precision was applied to all modules to test the limits of compression. This lossy compression may not fully preserve the high-precision information required for "Medical" or "Reference" tasks. This analysis suggests that a potential mitigation strategy, enabled by the CMDQ framework, would be the application of "module-specific bit-widths." The "Modality-Specific Module Partitioning Strategy") inherently supports such mixed-precision quantization. The observed performance decline appears primarily linked to modules responsible for processing fine-grained visual features. Consequently, the CMDQ framework provides a path for future work to explore assigning higher bit-widths (e.g 8-bit) specifically to the "vision encoding modules" and the vision-specific components within the "cross-modal processing modules". Other components, such as those processing language, could remain at 4-bit. It is hypothesized that this differential setup could mitigate the performance drop on high-sensitivity tasks, achieving a more optimal balance between accuracy and efficiency, which represents an important direction for future investigation.}

Compared to models with similar parameters but higher memory usage, such as Pixtral-12B and ShareGPT4V-13B, the quantized model shows a slight gap in overall performance but performs better in the History category, while reducing memory requirements by about 40\%. This advantage is particularly important in resource-constrained application scenarios, such as mobile devices or embedded systems.

Compared to small parameter models with similar memory usage (such as ShareGPT4V-7B, Mantis-8B-Idefics2, and LLaVA-LLaMA-3-8B), the quantized model shows a clear leading advantage in overall score, leading in the Education category and performing particularly well in the History category. Although some small models show outstanding performance in certain specific categories, such as Mantis-8B-Idefics2's 62.5 points in the Reference category, the quantized model still has a clear advantage in overall balance.

Combining results from both MMBench and OCRVQA datasets, it is evident that the quantization method not only significantly lowers the deployment threshold, enabling 19B-level large models to run efficiently on a single 4090 GPU, but also maintains limited performance loss while preserving various aspects of the model's capabilities. The quantization method significantly improves model deployment efficiency while maintaining advanced reasoning capabilities, creating broader application possibilities. Figure \ref{fig:radar_mmbench} and Figure \ref{fig:radar_ocrvqa} provide visual representations of these performance comparisons, highlighting the quantized model's balanced capabilities across different dimensions while significantly reducing memory requirements.

\subsection{Multi-Agent Assistance Framework Experiment}
{ To validate the practical functionality, robustness, and performance of the multi-agent assistance framework, a comprehensive series of experiments was designed. These experiments covered typical real-world test scenarios that visually impaired users might encounter in daily life, including various indoor settings such as offices, conference rooms, corridors, and desktop reading areas. The evaluation was multi-faceted and combined qualitative, quantitative, and system-level analyses.
	
First, we adopted a continuous scene capture approach (three consecutive frames) to qualitatively analyze the system's core workflows. This involved recording the system's actual outputs—such as scene descriptions, identified objects, detected behaviors, scene change determinations, and scene type classifications—to demonstrate how it processes and organizes visual information. This analysis also examined the system's ability to transition between scene types and the effectiveness of the RAG memory system in supplementing environmental information, including its performance under challenging illumination conditions and in specific failure-case scenarios.

Second, we conducted quantitative evaluations of the vectorized memory system's retrieval precision and stability. Using a pre-stored database and repeated queries, we analyzed the effectiveness of our dual-threshold filtering strategy. Finally, we measured key system-level performance benchmarks, focusing on interaction latency (time-to-first-speech) and hardware resource consumption (GPU/CPU memory and power), to verify the framework's efficiency and feasibility for real-world deployment. }
\subsubsection{Scene Classification Performance Evaluation}

To evaluate the performance of the multi-agent system in scene classification, experiments were designed with typical scene sequences that visually impaired users might encounter in daily life. This experiment focused on assessing the system's processing capabilities across three core scene types: text scenes (TEXT\_SCENE), obstacle scenes (OBSTACLE\_SCENE), and environmental description scenes (DESCRIPTION\_SCENE).

\begin{figure*}[htbp]
    \centering
    \includegraphics[width=0.9\textwidth]{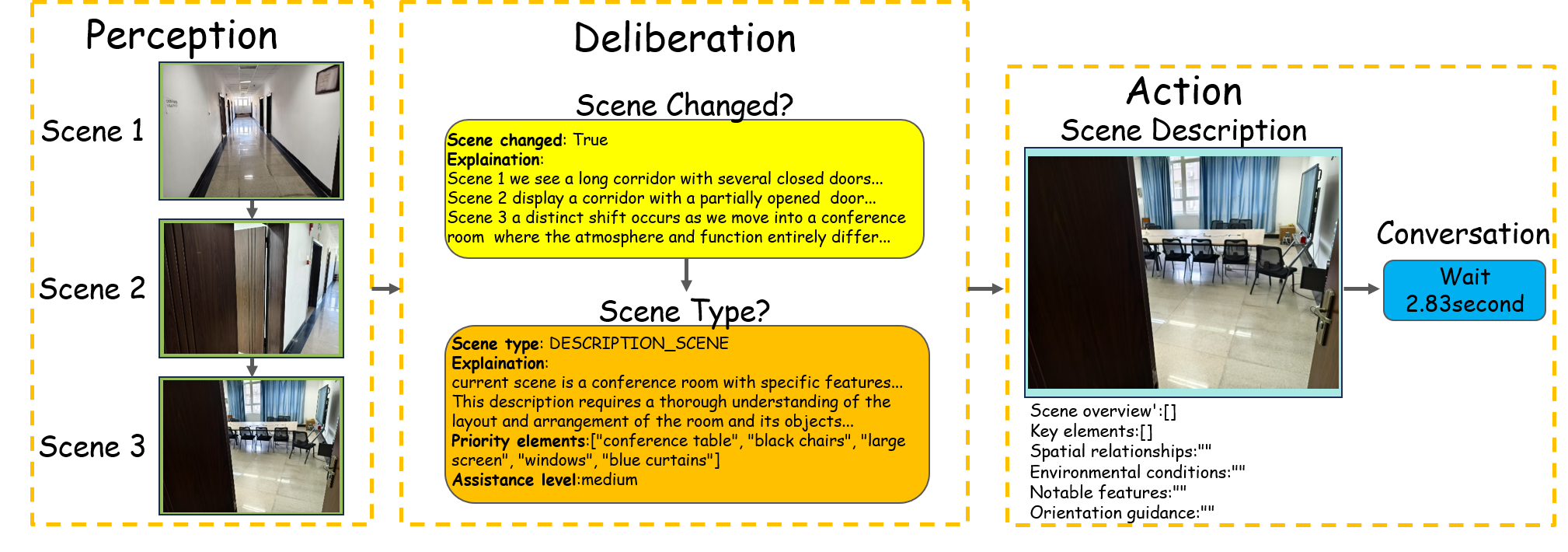} 
\caption{Example of environmental description scene processing workflow. The system successfully identified the environmental description type in the scene transition from corridor to conference room and extracted key layout information. The figure shows the complete perception-analysis-execution process, with detailed analysis results presented in Table~\ref{tab:description_scene_details}.}
 \label{fig:desc_scene}
\end{figure*}

\begin{figure*}[htbp]
    \centering
    \includegraphics[width=0.9\textwidth]{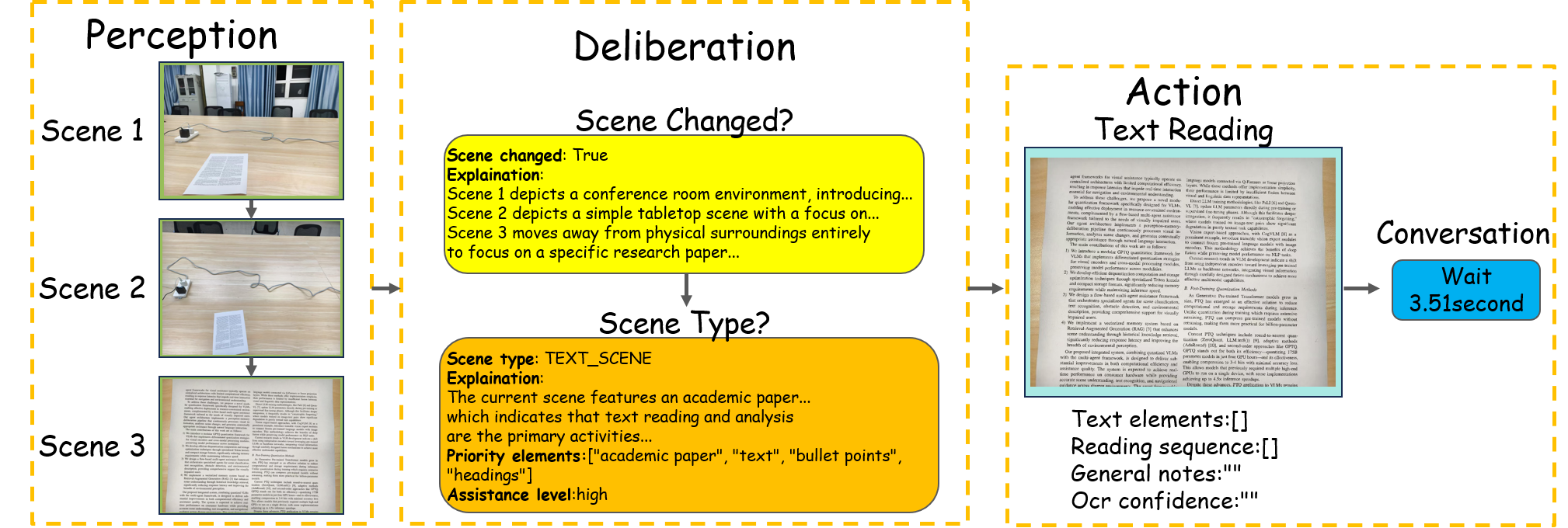} 
\caption{Example of text scene processing workflow. The system recognized the scene transition from conference room environment to academic paper close-up, correctly classified it as a text scene, and initiated the text recognition process. Complete text recognition results and confidence scores are shown in Table~\ref{tab:text_scene_details}.}
 \label{fig:text_scene}
\end{figure*}

\begin{figure*}[htbp]
    \centering
    \includegraphics[width=0.9\textwidth]{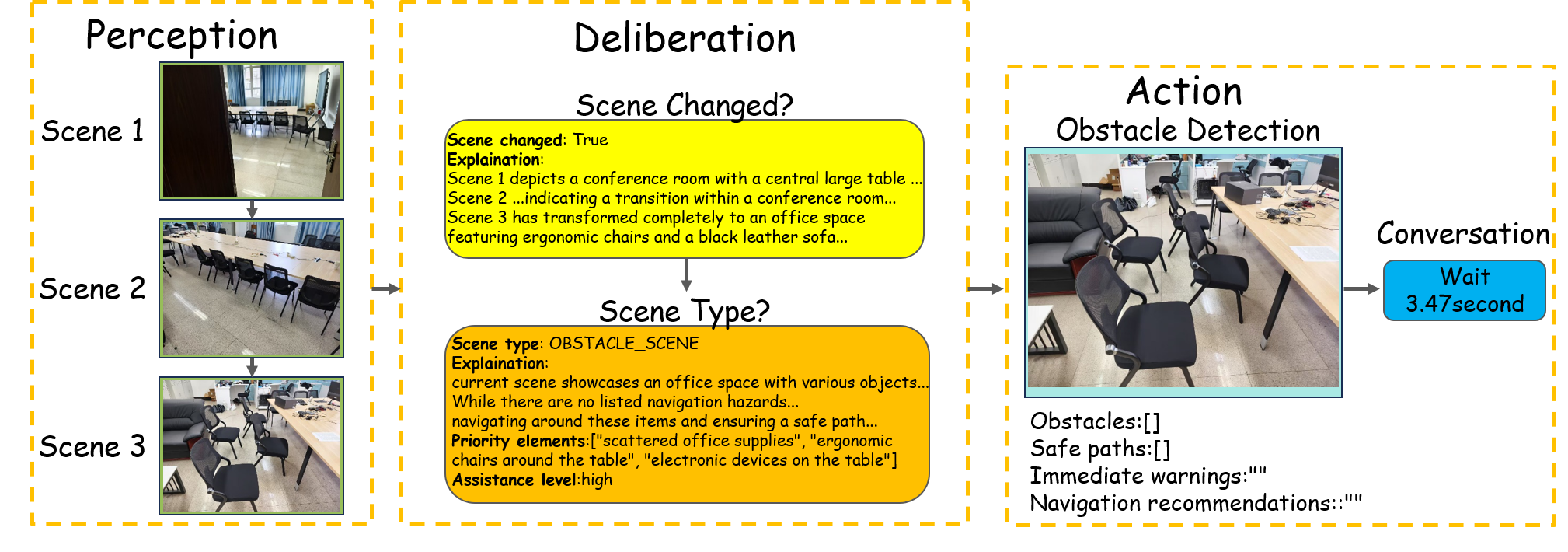} 
\caption{Example of obstacle scene processing workflow. The system identified potential navigation obstacles in the transition from conference room to office space and provided corresponding navigation recommendations. Obstacle analysis details and safe path planning are presented in Table~\ref{tab:obstacle_scene_details}.}
 \label{fig:obstacle_scene}
\end{figure*}

The experiment employed a continuous scene capture approach, with each test sequence containing three consecutive frames, simulating the process of visually impaired users moving through environments or shifting their attention. Multiple groups of scene sequences were collected in actual office environments, covering various indoor settings including corridors, conference rooms, office areas, and reading areas. The testing process followed the system's perception-analysis-interaction closed-loop design, with Figures~\ref{fig:desc_scene}, \ref{fig:text_scene}, and \ref{fig:obstacle_scene} illustrating the complete processing workflow for environmental description scenes, text scenes, and obstacle scenes, respectively.

As shown in Figure~\ref{fig:desc_scene}, in the environmental description scene sequence, the system captured the transition from a corridor to a conference room. The scene change detection module successfully identified significant environmental changes, and the scene type classification agent correctly categorized the environment as a DESCRIPTION\_SCENE type. The system subsequently identified key priority elements including "conference table," "black chairs," "large screen," "windows," and "blue curtains," and marked the scene assistance level as "medium."

Figure~\ref{fig:text_scene} demonstrates the system's processing of a text scene. In this sequence, the scene gradually transformed from a conference room environment to a close-up of an academic paper. The system correctly identified the scene change and accurately classified the final scene as a TEXT\_SCENE type. Priority elements were identified as "academic paper," "text," "bullet points," and "headings," with the assistance level marked as "high."

Figure~\ref{fig:obstacle_scene} presents the system's ability to process obstacle scenes. In this sequence, the scene transitioned from a conference room to an office space, and the system correctly classified the final scene as an OBSTACLE\_SCENE type. The system identified priority elements including "scattered office supplies," "ergonomic chairs around the table," and "electronic devices on the table," with the assistance level marked as "high."

Through testing multiple groups of scene sequences, the system demonstrated scene type classification capabilities across different environments, providing targeted analysis and assistance information for the three different scene types. The experimental results indicate that the scene classification system based on large vision-language models can accurately identify features of different scene types and make appropriate processing decisions. Particularly in complex environment transitions, the system exhibited stable classification performance, establishing a solid foundation for subsequent specialized processing.

\subsubsection{Vectorized Memory Retrieval Performance}

To evaluate the effectiveness of the vectorized memory system in practical applications, specific test scenarios were designed simulating situations where visually impaired users navigate familiar environments. The test environment included wooden tables, office items, and fixed features, with a focus on assessing how the system supplements the limitations of current perspective through historical memory.

\begin{figure*}[htbp]
    \centering
    \includegraphics[width=0.9\textwidth]{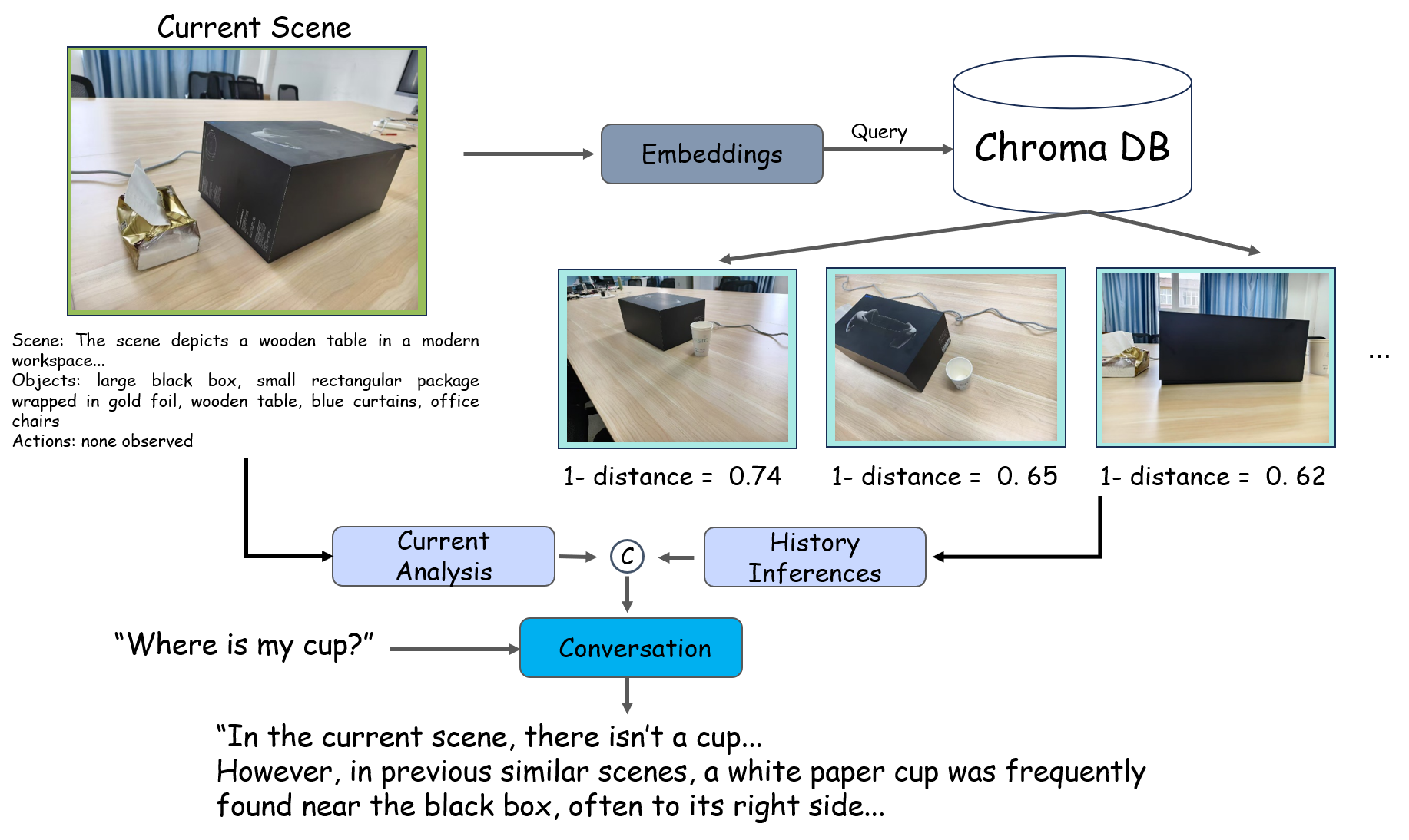} 
\caption{Performance evaluation of RAG-based memory retrieval. The figure shows the complete workflow of current scene analysis, vector retrieval, and historical scene integration. The system successfully utilized historical scene knowledge (three similar scenes on the right) to answer the location of objects not visible in the current scene, demonstrating cross-perspective reasoning capabilities. See Table \ref{tab:rag_memory_details} for detailed experimental results.}
 \label{fig:rag_result}
\end{figure*}

As shown in Figure \ref{fig:rag_result}, the test case demonstrates the system's processing capability for the current scene (containing a wooden table, large black box, gold-wrapped item, etc.). The system successfully retrieved three historical scenes with similarities of 0.74, 0.65, and 0.62, all of the "DESCRIPTION\_SCENE" type. Other scenes were excluded due to similarity scores below the minimum threshold, validating the effectiveness of the two-level threshold filtering strategy.

The key case in the testing demonstrated the system's environmental understanding capability when a user asked, "Where is my cup?" Although there was no cup in the current scene, the system accurately inferred based on item location information from historically similar scenes: "In previous similar scenes, a white paper cup was often located to the right side of the black box," providing valuable navigation guidance to the user. This result verified that the system could successfully overcome single-perspective cognitive limitations through the memory repository.

{The experimental results demonstrated that the vectorized memory system is capable of incrementally building a more thorough environmental awareness as it is used. By aggregating environmental data from various viewpoints, the system can offer spatial details that extend beyond the user's immediate field of vision , thereby assisting users in constructing more comprehensive mental models of their surroundings. This capability is particularly effective in static environments that are visited frequently, as the system's performance organically enhances with prolonged use without necessitating any further model training or fine-tuning.}

This environment understanding method enhanced by historical data provides visually impaired users with more comprehensive environmental information, compensating for the limitations of traditional single-perspective analysis methods. The cross-perspective reasoning ability observed in the tests demonstrates the practical value of this method in real assistance scenarios, especially for helping visually impaired users understand spatial information beyond their current perspective. Table \ref{tab:rag_memory_details} shows the detailed analysis results of a complete RAG memory retrieval experiment case.

{
	\begin{figure*}[htbp]
		\centering
		\includegraphics[width=0.9\textwidth]{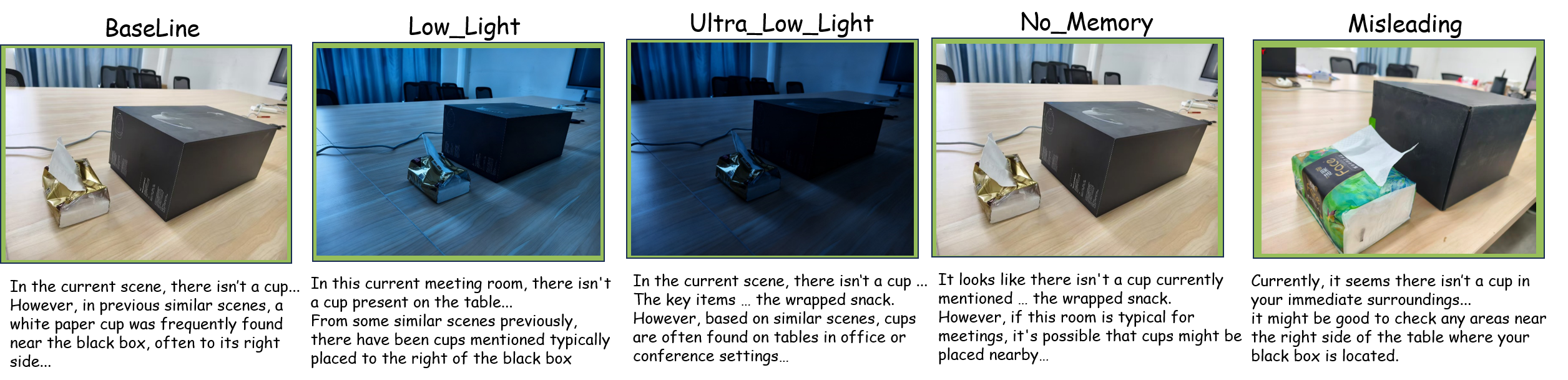}
		\caption{{Qualitative analysis of RAG system robustness and failure modes. (a) Standard condition (baseline). (b) Medium-low illumination. (c) Very-low illumination, where perception fails. (d) Input scene for the no-memory ablation test. (e) Misleading counterfactual scene with a tissue bag, which caused the RAG system to retrieve incorrect historical associations.}}
		\label{fig:ablation_robustness}
	\end{figure*}
	
	To further validate the necessity of the vectorized memory system, we conducted a critical ablation study. In this experiment, we retained the complete Multi-Agent System framework but disabled the memory retrieval module. Under this setting, the system could only rely on current single-frame perception. When faced with the same user query as in Figure \ref{fig:rag_result} ("Where is my cup?"), the system was unable to utilize any historical knowledge due to the absence of the cup in the current view and could only reply that it did not detect a cup. This forms a stark contrast to the successful inference using memory in Figure \ref{fig:rag_result}, strongly demonstrating the key role of the RAG module in providing contextual information beyond the current field of view and achieving cross-perspective understanding.
	
	Furthermore, we constructed a qualitative "memory-induced counterfactuals" failure case, as shown in Figure \ref{fig:ablation_robustness}(e). We established a scene with a layout highly similar to Figure \ref{fig:rag_result} but replaced the key "large black box" and "tissue bag" with a different model of black box and a tissue bag. Despite the change in key objects, the RAG system still retrieved the historical memory from Figure \ref{fig:rag_result} due to the high semantic similarity of the overall environment. Consequently, the system incorrectly generalized the old association to the new objects, replying to the user, "In previous similar scenes, a white paper cup was frequently found near the black box." This case exposes the limitations of the current implementation: the system relies on the semantic similarity of the overall scene during retrieval and cannot yet perform fine-grained verification of the specific identity of key objects, which may lead to misleading inferences in certain situations.
	
	Finally, we evaluated the system's robustness under varying illumination conditions, as shown in Figures \ref{fig:ablation_robustness}(b) and (c). We modified the original image from Figure \ref{fig:ablation_robustness}(a) using data augmentation to simulate different levels of low light. Experiments found that under medium-low illumination conditions (Figure \ref{fig:ablation_robustness}(b)), the Perception Agent could still successfully identify key elements such as the "large black box" and "wooden table," keeping the vectorized query effective and allowing the RAG system to correctly retrieve historical information and infer the cup's location. However, when the illumination conditions were further reduced (Figure \ref{fig:ablation_robustness}(c), simulating an extremely dark environment), the Perception Agent's performance dropped significantly, misidentifying the "large black box" as a "shadow" or "a bag." This erroneous perception result generated an invalid query vector, causing the RAG retrieval to recall incorrect memories or fail to find a match, ultimately preventing the system from providing correct assistive information. This indicates that the system's robustness is currently highly dependent on the accuracy of the upstream perception module under different lighting conditions.
}

{
\subsubsection{Quantitative Analysis and Ablation Study of Vectorized Memory}
\label{sec:quant_rag_ablation}

To quantitatively evaluate the effectiveness of our vectorized memory system, the necessity of our dual-threshold strategy, and the system's stability when handling similar inputs, we designed a set of retrieval and ablation experiments.

We first pre-stored a fixed memory database in ChromaDB consisting of 100 different indoor scene images. Among these 100 historical scenes, 5 were manually labeled as "Ground Truth" because they all contained the specific "cup" and "black box" association (highly relevant to the scene in Figure \ref{fig:rag_result}).

To simulate the query fluctuations the system might encounter in real-world operation, we used the current scene shown in Figure \ref{fig:rag_result} as a single input and ran our multi-agent framework N=200 consecutive times. Due to subtle descriptive differences generated by the framework's perception and analysis modules in each run, these 200 runs produced 200 slightly different "scene analysis" results. We used these analysis results directly as the query inputs, $E(S_{curr})$, to test the system's average retrieval performance.

We set the system to retrieve the k=5 most similar results. These k=5 items are not used directly; they first pass through a mandatory "scene label" filter. Any item with a non-matching scene label is immediately discarded. Only the remaining items are passed to the similarity threshold stage (governed by $\tau_{high}$ and $\tau_{low}$). The data presented in our results are the average outcomes from these 200 queries, reflecting the performance of this full, two-stage filtering pipeline.

Second, only the items that pass this initial label check are then evaluated against the dual-similarity thresholds. The data presented in our results are the average outcomes from these 200 queries, reflecting this full pipeline.

We evaluated four different threshold configurations to verify the roles of the high threshold $\tau_{high}$ (for direct reuse) and the low threshold $\tau_{low}$ (for reference filtering):

\textbf{Standard Configuration (Baseline):} Used our system's default values ($\tau_{high}=0.80$, $\tau_{low}=0.55$). This uses both scene label filtering and similarity thresholding.

\textbf{High-Threshold Ablation (No-Reuse):} $\tau_{high}$ was set to 1.0, making it impossible to trigger "direct reuse." All retrieved results that pass both filters are used only as "reference."

\textbf{Low-Threshold Ablation (No-Filter):} $\tau_{low}$ was set to 0.0, disabling \textit{only} the similarity filtering. The mandatory "scene label" filtering still applies. All results passing the scene label check are used for subsequent processing, regardless of score.

\textbf{Dual Ablation:} $\tau_{high}=1.0$ and $\tau_{low}=0.0$ were set simultaneously (disabling the entire similarity threshold stage, but not the scene label filter).

\begin{table*}[htbp] % 使用 table* 来实现跨双栏
	\centering
	\caption{{Summary of Ablation Study Results (Average over N=200 runs)}}
	\label{tab:ablation_results}
	\resizebox{0.9\textwidth}{!}{ % 使用 0.9\textwidth，与您的示例一致
		\begin{threeparttable}
			\begin{tabular}{@{}lccccc@{}} % 6列，第1列左对齐，其余居中
				\toprule[2pt]
				\textbf{Scenario (Ablation)} & \textbf{$\tau_{high}$} & \textbf{$\tau_{low}$} & \textbf{Avg. P@k} & \textbf{Avg. High-Match} & \textbf{Avg. False Positives} \\
				\midrule
				Standard Configuration & 0.80 & 0.55 & 0.8082 & 0.0500 & 0.65 \\
				High-Thresh. Ablation (No-Reuse) & 1.00 & 0.55 & 0.8082 & 0.0000 & 0.65 \\
				Low-Thresh. Ablation (No-Filter) & 0.80 & 0.00 & 0.8548 & 0.0500 & 0.72 \\
				Dual Ablation & 1.00 & 0.00 & 0.8548 & 0.0000 & 0.72 \\
				\bottomrule[2pt]
			\end{tabular}
			\begin{tablenotes}
				\small
				\item \textbf{P@k (Precision@k):} The average ratio of True Positives (TPs) among the total k=5 retrieved items.
				\item \textbf{High-Match Success:} The average rate at which a retrieved item surpassed $\tau_{high}$ and was also a True Positive.
				\item \textbf{Avg. False Positives:} The average count of items (out of k=5) that were retrieved but did not match the Ground Truth rule.
			\end{tablenotes}
		\end{threeparttable}
	}
\end{table*}

The experimental results, summarized in \textbf{Table \ref{tab:ablation_results}}, clearly demonstrated the effectiveness of the dual-threshold strategy. We primarily focused on the average retrieval precision (Avg. P@k), high-similarity match success (High-Match Success), and average false positives (Avg. False Positives).

By comparing the "Standard Configuration" and "High-Threshold Ablation," we found that P@k (0.8082) and FPs (0.65) remained unchanged. The only difference was that "High-Match Success" dropped from 0.05 to 0.00. This proves that the $\tau_{high}$ threshold (reuse mechanism) is a pure optimization strategy. It successfully identifies and reuses high-similarity scenes without impacting core retrieval quality.

Meanwhile, the $\tau_{low}$ threshold played a critical role as a secondary noise filter. Our system's primary filter (the "scene label" check) is always active in all configurations, removing categorically irrelevant items first. The "Low-Threshold Ablation" case ($\tau_{low}=0.0$) thus represents a baseline where only these scene-label-matched items are considered.

When comparing "Standard Configuration" ($\tau_{low}=0.55$) and "Low-Threshold Ablation," we isolate the effect of the \textit{additional} similarity threshold. We observed a typical precision/recall trade-off: the average false positives (FPs) increased significantly from 0.65 to 0.72 (an increase of approx. 10.8\%) when this secondary $\tau_{low}$ filter was removed. This demonstrates that the $\tau_{low}=0.55$ threshold successfully filtered out a large amount of irrelevant, low-similarity memory noise \textit{that had already passed the initial scene label check}. Although P@k also slightly decreased (from 0.8548 to 0.8082), we believe that sacrificing minimal precision to achieve a significant reduction in false positives is a worthwhile trade-off. This reduces the risk of the system being misled by irrelevant memories from the correct scene category.

In summary, our "Standard Configuration" achieves the optimal balance between retrieval quality, noise suppression, and computational efficiency.
}

{
\subsubsection{System Performance Evaluation}
}
To quantify the effectiveness of the streaming speech processing approach in reducing interaction latency, comparative experiments were conducted measuring the time from model response generation to the first audible speech output. This metric is particularly critical for visually impaired users who rely entirely on audio feedback for environmental understanding. As shown in Tables \ref{tab:obstacle_scene_details}, \ref{tab:text_scene_details} and \ref{tab:description_scene_details}, the system achieved consistent response times between 2.83 and 3.52 seconds across different scene types. These measurements represent the complete processing pipeline from scene analysis to the first sentence being spoken, demonstrating the system's ability to provide timely feedback in real-world scenarios.
{To further justify the system's suitability for embedded or mobile hardware, we analyzed the resource footprint across its core operational stages, as requested by the reviewer. We profiled two primary workflows: (1) the vision-informed workflow, which encompasses the stages of visual \textit{perception}, \textit{LLM}-based generation, and \textit{TTS}, and (2) the text-informed workflow, which focuses on the \textit{deliberation} and \textit{retrieval} stages. The resource consumption metrics, measured on consumer-grade hardware, are detailed in Table \ref{tab:resource_consumption}.}

{The results in Table \ref{tab:resource_consumption} provide compelling evidence of the system's efficiency for assistive settings. The primary finding is the exceptional memory management enabled by our CMDQ framework. Peak GPU memory is constrained to under 11.6 GB, even during the more demanding vision-informed workflow. This significant reduction from the original 38GB model size is pivotal for deployment on widely available consumer hardware, a critical requirement for practical assistive technology. Furthermore, the moderate power draw, averaging around 130W during active inference, establishes a vital baseline for estimating the energy envelope for future battery-powered devices, directly addressing the operating concerns for mobile hardware.}
The experimental results indicate that the streaming speech synthesis approach significantly reduces perceived latency compared to traditional methods. In conventional text-to-speech pipelines without streaming capabilities, the system would need to complete the entire response generation before beginning speech synthesis, resulting in interaction delays exceeding 30 seconds for comprehensive scene descriptions. This substantial delay would severely impact usability for visually impaired individuals, particularly in dynamic environments requiring quick decisions.
The sentence-level chunking mechanism implemented in the system enables concurrent processing of text generation and speech synthesis, providing immediate auditory feedback while the remainder of the response is still being generated. This approach maintains a balance between response comprehensiveness and interaction fluidity, ensuring that users receive critical information without prohibitive delays. By focusing on optimizing the interaction layer, a system was created that delivers information in a timely manner without forcing users to wait for complete response generation{, making it highly suitable for real-world deployment scenarios with visually impaired users}.

\begin{table*}[htbp] % 使用 table* 来实现跨双栏
	\centering
	\caption{{System Resource Footprint Across Core Workflows}}
	\label{tab:resource_consumption}
	\resizebox{0.9\textwidth}{!}{ % 调整宽度以适应页面
		\begin{threeparttable}
			\begin{tabular}{@{}lcccc@{}} % 5列，第1列左对齐，其余居中
				\toprule[2pt]
				\textbf{Assistive Workflow} & \textbf{Core Stages Involved} & \textbf{Peak GPU Memory (MB)} & \textbf{Peak CPU Memory (MB)} & \textbf{Avg. GPU Power (W)} \\
				\midrule
				Vision-Informed Workflow & Visual \textit{Perception}, \textit{LLM} Generation, \textit{TTS} & 11,584 & 11,023 & 129.9 \\
				Text-Informed Workflow & Text-based \textit{Deliberation} , \textit{Retrieval} & 11,580 & 11,021 & 126.1 \\
				\bottomrule[2pt]
			\end{tabular}
			\begin{tablenotes}
				\small
				\item Measurements were conducted on a high-performance consumer-grade GPU to establish a performance baseline for assistive technology applications.
			\end{tablenotes}
		\end{threeparttable}
	}
\end{table*}

{
\section{Discussion}\label{Sec_5}

This section provides a comprehensive analysis of the study's findings and the framework's architecture. We first present the primary experimental results, detailing the achievements in both model optimization and application-level performance. Following this, we examine the key methodological advantages that enabled these outcomes, as well as the inherent limitations and methodological considerations of the current framework. Finally, we conclude by discussing the broader insights and implications derived from this research.

\subsection{Research Results}
The experimental results of this study demonstrate significant achievements in both model efficiency and application-level performance. At the model level, our proposed framework successfully reduced the memory footprint of a 19B-parameter Vision-Language Model (VLM) from approximately 38GB to 11.3GB. This substantial compression was accomplished with minimal impact on performance, as evidenced by a performance degradation of only 2.05\% on the MMBench benchmark and a score reduction of just 1.2 points on the OCR-VQA task. This result confirms the feasibility of running large-scale VLMs on consumer-grade hardware without significant functional compromise.

At the application level, our framework proved its functional efficacy through its sophisticated architecture. The flow-based multi-agent system successfully orchestrated complex workflows, correctly classifying dynamic scenes and dispatching specialized agents for analysis. Furthermore, the system demonstrated its ability to perform advanced reasoning by leveraging its Retrieval-Augmented Generation (RAG) capabilities. It successfully integrated historical scene data to provide cross-perspective understanding, offering assistance that goes beyond immediate visual perception. This entire process, from perception to complex reasoning, was executed with high efficiency, resulting in an end-to-end response time in the range of 2.83 to 3.52 seconds.

\subsection{Methodological Advantages}
The successful results are attributable to several key methodological advantages. The primary innovation is CMDQ framework, which addresses the architectural heterogeneity of VLMs. Unlike generic quantization methods, its differentiated, modular strategy respects the varying sensitivities of the visual encoder and cross-modal modules, leading to a more optimal balance between compression and performance preservation. This is complemented by efficient dequantization and storage optimization techniques that ensure high inference speed post-compression.

The core advantage of our multi-agent framework lies in the synergy between its components. The scene-aware vectorized memory system, powered by RAG, overcomes the inherent limitations of single-perspective analysis to enable a more holistic environmental understanding. This advanced capability is managed and made actionable by a flow-based collaborative architecture that systematically orchestrates diverse tasks. The integration of a speech streaming pipeline is a crucial final step, ensuring that the rich, contextual information generated by the system is delivered to the user with minimal delay, thus maximizing its practical utility in real-world scenarios.

\subsection{Limitations and Methodological Considerations}
While the framework demonstrates considerable strengths, a comprehensive understanding requires examining its limitations and the implications of its methodological choices.

Methodologically, the contribution centered on the CMDQ \textit{framework} for managing cross-modal sensitivities, which adapted established quantization principles rather than designing novel underlying \textit{algorithms}. This focus leaves potential algorithm-level optimizations, tailored to VLM data distributions, as a promising avenue for future work. Architecturally, the framework lacks a robust self-validation or correction mechanism. This introduces a dual risk: the flow-based system is susceptible to error propagation from upstream faults, and the RAG-based memory could generate "memory-induced counterfactuals" if retrieved data is subtly inapplicable to the current context.

The study's generalizability is also constrained by several factors. System evaluation was conducted using simulated scenarios due to the absence of a large-scale public benchmark for diverse, real-world assistive tasks. Validation was primarily limited to controlled indoor environments, and the model calibration utilized a general-purpose dataset. The system's robustness in dynamic outdoor settings, or its performance with task-specific calibration data, has yet to be systematically evaluated. Finally, the framework relies solely on visual information, inherently limiting its ability to perceive precise spatial metrics like distance and depth, which are critical for fine-grained navigation.

\subsection{Insights and Implications}
Beyond the quantitative results, the research process itself yielded several critical insights that fundamentally shaped the final architecture and design philosophy of the framework. These lessons concern the unique challenges of optimizing multimodal models and the practical requirements for user-centric assistive systems.

A primary insight gained during this research is the critical need for a differentiated approach to multimodal model optimization. Initial attempts using generic quantization schemes yielded suboptimal results, which led to the realization that components processing different modalities—such as the visual encoder and cross-modal fusion modules—exhibit distinct sensitivities to quantization. This fundamental understanding directly motivated the development of the CMDQ framework as a more nuanced and effective solution.

It also became evident that even a perfect description of the immediate visual field is insufficient to meet the holistic needs of a user in an assistive context. A key revelation was that truly intelligent assistance must transcend single-frame perception to build a spatio-temporally coherent understanding of an environment. The integration of a vectorized memory system to enable "cross-perspective understanding" was therefore identified as a decisive step in evolving the system from a descriptive 'tool' into a context-aware 'companion'.

Finally, the profound impact of interaction latency on user experience was a crucial lesson. Early prototypes, while functionally capable, suffered from slow response times that diminished their practical utility and could lead to user frustration. This underscored that computational efficiency must ultimately serve the end-user experience. Consequently, significant effort was directed toward optimizing dequantization kernels and implementing speech streaming, which successfully reduced the perceived response time from potentially over 30 seconds to under 4 seconds, fundamentally improving the system's real-world usability.
}
\section{Conclusion}\label{Sec_6}
This study has successfully addressed technical and practical challenges in vision-language model deployment for assistive technology. The proposed modular quantization framework effectively compressed a 19B parameter model from 38GB to 11.3GB with minimal performance impact, demonstrating only a 2.05\% accuracy reduction on MMBench and maintaining 63.7 accuracy on OCR-VQA. The scene-aware vectorized memory multi-agent system has provided comprehensive assistance for visually impaired individuals by integrating scene classification, vectorized memory, and multimodal interaction. The system achieves persistent scene memory storage and efficient retrieval, enabling environmental understanding beyond the immediate view, with response latency dramatically reduced to 2.83-3.52 seconds from traditional methods' 30+ seconds. 

{
These technological innovations advance computational efficiency and assistive technology, transforming complex AI systems into practical tools that meaningfully improve visually impaired individuals' daily lives. Building on this foundation, future work will proceed in several key directions. We aim to further refine the quantization framework by exploring novel, algorithm-level optimizations tailored to VLM data distributions and implementing flexible, mixed-precision strategies. Concurrently, we will enhance the multi-agent system's robustness against error propagation and integrate additional sensors, such as depth cameras or LiDAR, to provide more precise navigational guidance. Finally, we will validate these advancements by expanding evaluation from controlled scenarios to more dynamic, real-world outdoor environments and public benchmarks, establishing a foundation for a more inclusive technological future.}

\appendix
\section{Detailed Scene Analysis Results}
In this appendix, detailed analysis results from the experimental evaluation of the multi-agent assistance framework are provided. Tables \ref{tab:obstacle_scene_details}, \ref{tab:text_scene_details}, and \ref{tab:description_scene_details} present the comprehensive processing results for obstacle detection scenes, text recognition scenes, and environmental description scenes, respectively. Additionally, Table \ref{tab:rag_memory_details} demonstrates the complete retrieval and integration workflow of the RAG-based memory system.
\onecolumn
\begin{longtable}{p{2.5cm}p{12cm}}
\caption{Detailed Analysis Results of Obstacle Scene Processing\label{tab:obstacle_scene_details}}\\
\toprule
\textbf{Analysis Dimension} & \textbf{Results} \\
\midrule
\multicolumn{2}{c}{\textbf{Scene Information}} \\
\midrule
\endfirsthead

\multicolumn{2}{c}%
{{\bfseries Table \thetable{} -- continued from previous page}} \\
\toprule
\textbf{Analysis Dimension} & \textbf{Results} \\
\midrule
\endhead

\midrule
\multicolumn{2}{r}{{Continued on next page}} \\
\bottomrule
\endfoot

\bottomrule
\endlastfoot

Initial Scene (Scene 1) & "The image depicts a conference room viewed from an open door. The room has a large table set up in the center, surrounded by several black chairs. Blue curtains are drawn over the windows, allowing some natural light to enter the space. A large screen is positioned at one end of the room, and there are wires visible on the floor near the table." \\
& Identified Objects: ["conference table", "black chairs", "blue curtains", "large screen", "wires on the floor"] \\
& Actions: ["door being opened"] \\
\midrule
Transition Scene (Scene 2) & "A modern conference room with a long wooden table set in the middle, surrounded by black mesh chairs. The atmosphere is clean and organized, with a few electronic gadgets and office supplies scattered on the table. Large windows are covered with blue curtains, allowing some natural light to enter the room." \\
& Identified Objects: ["long wooden table", "black mesh chairs", "electronic gadgets", "office supplies", "blue curtains"] \\
& Actions: [] \\
\midrule
Target Scene (Scene 3) & "The scene depicts an office space featuring several ergonomic chairs arranged around a large wooden table. In the background, there is a black leather sofa and storage cabinets, along with various office supplies scattered around the room. The floor appears to be polished, and there are some electronic devices on the table." \\
& Identified Objects: ["ergonomic chairs", "wooden table", "black leather sofa", "storage cabinets", "computer monitor", "office supplies", "cables"] \\
& Actions: ["chairs are arranged around the table", "supplies are scattered on the table"] \\
\midrule
Scene Change Analysis & 'scene\_changed': 'True' \\
& 'explanation': 'The analysis of the three Scenes suggests a significant scene change has occurred from the initial to the final location. Scene 1 depicts a conference room with a central large table and basic furnishings, characterized by a corporate environment. In Scene 2, there is a long wooden table and an organized setup indicating a transition within a conference room, but it retains elements similar to Scene 1, fostering continuity. By Scene 3, however, the scene has transformed completely to an office space featuring ergonomic chairs and a black leather sofa, indicating a different purpose and environment as it reflects a more relaxed office atmosphere, as opposed to a formal conference room. The presence of different furniture and office supplies further solidifies this location change, suggesting the transition is more than just a reorganization of the previous setting, but rather a move to an entirely different space designed for casual use rather than meetings.' \\
\midrule
Scene Type Analysis & "scene\_type": "OBSTACLE\_SCENE" \\
& "explanation": "The current scene showcases an office space with various objects such as ergonomic chairs, a large wooden table, a black leather sofa, storage cabinets, and scattered office supplies. While there are no listed navigation hazards, the presence of scattered supplies and electronic devices on the table indicates potential collision risks. The user might need assistance navigating around these items and ensuring a safe path to exit or move throughout the room. Given the arrangement of chairs and other elements, careful navigation is required, prioritizing user safety." \\
& "priority\_elements": ["scattered office supplies", "ergonomic chairs around the table", "electronic devices on the table"] \\
& "assistance\_level": "high" \\
\midrule
\multicolumn{2}{c}{\textbf{Obstacle Detection Results}} \\
\midrule
Obstacle 1 & Type: chairs \\
& Location: in front of the workspace table \\
& Distance: near the main walking path \\
& Risk Level: medium \\
& Avoidance Suggestion: walk around the chairs from either side to prevent tripping. \\
\midrule
Obstacle 2 & Type: couch \\
& Location: against the wall \\
& Distance: close to the pathway \\
& Risk Level: low \\
& Avoidance Suggestion: maintain a safe distance while passing. \\
\midrule
Obstacle 3 & Type: wires and cables \\
& Location: on the desk near the computer setup \\
& Distance: above the floor, could lead to tripping if walked into \\
& Risk Level: medium \\
& Avoidance Suggestion: navigate around the desk area carefully. \\
\midrule
Obstacle 4 & Type: box \\
& Location: on the table \\
& Distance: within the workspace area \\
& Risk Level: low \\
& Avoidance Suggestion: ensure not to bump into it while maneuvering. \\
\midrule
\multicolumn{2}{c}{\textbf{Safe Paths}} \\
\midrule
Safe Path 1 & Direction: around chairs \\
& Distance: 2 meters \\
& Description: space between the chairs to walk through \\
& Clearance: 1 meter wide \\
\midrule
Safe Path 2 & Direction: towards the wall \\
& Distance: 3 meters \\
& Description: path along the wall past the couch \\
& Clearance: 1.5 meters \\
\midrule
\multicolumn{2}{c}{\textbf{Navigation Guidance}} \\
\midrule
Immediate Warnings & "Be cautious of the chair arrangement as they can easily cause tripping." \\
\midrule
Navigation Recommendations & "Always keep an eye on the path in front and avoid moving too quickly around obstacles." \\
\midrule
\multicolumn{2}{c}{\textbf{System Performance}} \\
\midrule
System Response Time & 3.47 seconds \\
\end{longtable}

\onecolumn
\begin{longtable}{p{2.5cm}p{12cm}}
\caption{Detailed Analysis Results of Text Scene Processing\label{tab:text_scene_details}}\\
\toprule
\textbf{Analysis Dimension} & \textbf{Results} \\
\midrule
\multicolumn{2}{c}{\textbf{Scene Information}} \\
\midrule
\endfirsthead

\multicolumn{2}{c}%
{{\bfseries Table \thetable{} -- continued from previous page}} \\
\toprule
\textbf{Analysis Dimension} & \textbf{Results} \\
\midrule
\endhead

\midrule
\multicolumn{2}{r}{{Continued on next page}} \\
\bottomrule
\endfoot

\bottomrule
\endlastfoot

Initial Scene (Scene 1) & "A wooden table with a light-colored finish is set in a room with no visible decoration. On the table, there is an electrical outlet strip with a power cord connected to a device. There is also a printed document placed on the table, partially visible." \\
& Identified Objects: ["wooden table", "electrical outlet strip", "power cord", "printed document"] \\
& Actions: [] \\
\midrule
Transition Scene (Scene 2) & "The scene depicts a quiet, well-lit meeting room featuring a long wooden table with several chairs around it. The walls are adorned with blue curtains, and there is a storage cabinet to one side, alongside an air conditioning unit. The ambiance suggests a space meant for discussions or meetings." \\
& Identified Objects: ["long wooden table", "several chairs", "blue curtains", "storage cabinet", "air conditioning unit", "power strip", "sheet of paper"] \\
& Actions: ["none observed; the scene appears to be static"] \\
\midrule
Target Scene (Scene 3) & "The image appears to be a page of printed text, possibly an academic paper or technical document discussing frameworks for visual assistance in resource-constrained environments. The text is densely packed with information related to quantization methods and modular frameworks designed for visually impaired users." \\
& Identified Objects: ["printed text", "academic paper"] \\
& Actions: ["reading", "analyzing"] \\
\midrule
Scene Change Analysis & 'scene\_changed': 'True' \\
& 'explanation': 'The analysis of the three scenes indicates a significant change in the scene and location. Scene 1 depicts a simple tabletop scene with a focus on a power strip, power adapter, and a printed sheet of paper, suggesting a personal workspace or a casual setting. In contrast, Scene 2 transitions into a conference room environment, introducing a larger setting with a wooden table and chairs, indicating a communal or formal atmosphere. This transition highlights a more organized and professional context, as seen with elements like curtains and storage cabinets. Finally, Scene 3 moves away from physical surroundings entirely to focus on a specific study paper discussing advanced frameworks. This shift indicates not just a change in physical location—moving from a casual workspace to a conference room, then to an academic focus—but also a transition in subject matter, from physical objects to intellectual content. Therefore, these sequential Scenes represent a clear progression through different scenes, showcasing a significant location and contextual change.' \\
\midrule
Scene Type Analysis & "scene\_type": "TEXT\_SCENE" \\
& "explanation": "The current scene features an academic paper discussing visual language models and methodologies, which indicates that text reading and analysis are the primary activities taking place. The presence of bullet points, headings, and the academic nature of the content confirms that this scene is focused on processing written information." \\
& "priority\_elements": ["academic paper", "text", "bullet points", "headings"] \\
& "assistance\_level": "high" \\
\midrule
\multicolumn{2}{c}{\textbf{Text Element Analysis}} \\
\midrule
Text Element 1 & Content: "agent frameworks for visual assistance typically operate on centralized architectures with limited computational efficiency, resulting in response latencies that impede real-time interaction essential for navigation and environmental understanding." \\
& Type: document \\
& Location: top half of the page \\
& Readability: high \\
& Importance: high \\
\midrule
Text Element 2 & Content: "To address these challenges, we propose a novel modular quantization framework specifically designed for VLMs, enabling effective deployment in resource-constrained environments, complemented by a flow-based multi-agent assistance framework tailored to the needs of visually impaired users." \\
& Type: document \\
& Location: beneath the introduction of agent frameworks \\
& Readability: high \\
& Importance: high \\
\midrule
Text Element 3 & Content: "Our agent architecture implements a perception-memory formation, analyzes scene changes, and generates contextually appropriate assistance through natural language interaction." \\
& Type: document \\
& Location: continuation of previous text \\
& Readability: high \\
& Importance: high \\
\midrule
Text Element 4 & Content: "The main contributions of this work are as follows:" \\
& Type: document \\
& Location: section header \\
& Readability: high \\
& Importance: high \\
\midrule
Text Element 5 & Content: "1) We introduce a modular GPTQ quantization framework for VLMs that implements differentiated quantization strategies for visual encoders and cross-modal processing modules, preserving model performance across modalities." \\
& Type: document \\
& Location: itemized contribution list \\
& Readability: high \\
& Importance: high \\
\midrule
Text Element 6 & Content: "2) We develop efficient dequantization computation and storage optimization techniques through specialized Triton kernels and compact storage formats, significantly reducing memory requirements while maintaining inference speed." \\
& Type: document \\
& Location: continuation of the contribution list \\
& Readability: high \\
& Importance: high \\
\midrule
Text Element 7 & Content: "3) We design a flow-based multi-agent assistance framework that orchestrates specialized agents for scene classification, text recognition, obstacle detection, and environmental description, providing comprehensive support for visually impaired users." \\
& Type: document \\
& Location: continuation of the contribution list \\
& Readability: high \\
& Importance: high \\
\midrule
Text Element 8 & Content: "4) We implement a vectorized memory system based on Retrieval-Augmented Generation (RAG) that enhances scene understanding through historical knowledge retrieval, significantly reducing response latency and improving the breadth of environmental perception." \\
& Type: document \\
& Location: continuation of the contribution list \\
& Readability: high \\
& Importance: high \\
\midrule
Text Element 9 & Content: "Our proposed integrated system, combining quantized VLMs with the multi-agent framework, is designed to deliver substantial improvements in both computational efficiency and assistance quality." \\
& Type: document \\
& Location: conclusion of the contributions section \\
& Readability: high \\
& Importance: high \\
\midrule
Text Element 10 & Content: "Current PTQ techniques include round-to-nearest quantization (ZeroQuant, LLM.int8()), adaptive methods (AdaRound), and second-order approaches like GPTQ." \\
& Type: document \\
& Location: discussion on PTQ techniques \\
& Readability: high \\
& Importance: high \\
\midrule
Text Element 11 & Content: "GPTQ stands out for both its efficiency—quantizing 175B parameter models in just four GPU hours—and its effectiveness, enabling compression to 3-4 bits with minimal accuracy loss." \\
& Type: document \\
& Location: final thoughts on current techniques \\
& Readability: high \\
& Importance: high \\
\midrule
\multicolumn{2}{c}{\textbf{Overall Analysis}} \\
\midrule
Reading Sequence & ['0', '1', '2', '3', '4', '5', '6', '7', '8', '9'] \\
\midrule
General Notes & 'The text appears to focus on an advanced modular framework for visual language models (VLMs) and their application to enhance visual assistance for users, particularly in demanding environments. The document is well-structured, with clear headings and contributions outlined in a list format, making it easy to follow.' \\
\midrule
OCR Confidence & high \\
\midrule
System Response Time & 3.52 seconds \\
\end{longtable}

\onecolumn
\begin{longtable}{p{2.5cm}p{12cm}}
\caption{Detailed Analysis Results of Description Scene Processing\label{tab:description_scene_details}}\\
\toprule
\textbf{Analysis Dimension} & \textbf{Results} \\
\midrule
\multicolumn{2}{c}{\textbf{Scene Information}} \\
\midrule
\endfirsthead

\multicolumn{2}{c}%
{{\bfseries Table \thetable{} -- continued from previous page}} \\
\toprule
\textbf{Analysis Dimension} & \textbf{Results} \\
\midrule
\endhead

\midrule
\multicolumn{2}{r}{{Continued on next page}} \\
\bottomrule
\endfoot

\bottomrule
\endlastfoot

Initial Scene (Scene 1) & "A long, well-lit corridor with white walls and a polished floor. The corridor features several closed doors on both sides, and at the far end, there is a set of double doors that lead to an area with natural light. The floor reflects the lights above, enhancing the spacious feel of the corridor. A sign is visible on the left wall, providing information." \\
& Identified Objects: ["closed doors", "double doors at the end", "sign on the wall", "ceiling lights"] \\
& Actions: [] \\
\midrule
Transition Scene (Scene 2) & "The scene features a corridor with a partially opened wooden door on the left side. The door is designed with a dark wood finish, and it folds back to reveal a narrow opening. The walls are painted a light color, and there is a mix of wood and tiled flooring. In the corridor, there are several doorways visible along the right side, with a green exit sign and a wall clock in the background." \\
& Identified Objects: ["wooden door", "door frame", "corridor walls", "exit sign", "wall clock", "floor tiles", "doorways"] \\
& Actions: ["the door is partially opened"] \\
\midrule
Target Scene (Scene 3) & "A modern conference room with a long wooden table set in the middle, surrounded by black mesh chairs. The atmosphere is clean and organized, with a few electronic gadgets and office supplies scattered on the table. Large windows are covered with blue curtains, allowing some natural light to enter the room." \\
& Identified Objects: ["long wooden table", "black mesh chairs", "electronic gadgets", "office supplies", "blue curtains"] \\
& Actions: [] \\
\midrule
Scene Change Analysis & 'scene\_changed': 'True' \\
& 'explanation': "The analysis of the three Scenes indicates a significant scene change from the initial to the final location. In Scene 1, we see a long corridor with several closed doors and double doors leading to an area of natural light, establishing a specific environment focused on hallway structure. By Scene 2, the scene evolves to display a corridor with a partially opened wooden door, which suggests further access to another area. The presence of an exit sign and clock in this Scene adds elements of time and navigational clues, indicating a transitional space. Finally, in Scene 3, a distinct shift occurs as we move into a conference room where the atmosphere and function entirely differ, characterized by natural light from large windows, a long table, and seating arrangements. This room is designed for meetings, offering a stark contrast to the corridor's function as merely a passageway. Overall, the series illustrates a clear movement from one location (the corridor) to an entirely different one (the conference room), confirming a significant scene change." \\
\midrule
Scene Type Analysis & "scene\_type": "DESCRIPTION\_SCENE" \\
& "explanation": "The current scene is a conference room with specific features such as large windows with blue curtains, polished flooring reflecting the interior, a long table in the center with black chairs surrounding it, and a large screen on the right. This description requires a thorough understanding of the layout and arrangement of the room and its objects for effective navigation and orientation within the space, thus fitting it into the description category." \\
& "priority\_elements": ["conference table", "black chairs", "large screen", "windows", "blue curtains"] \\
& "assistance\_level": "medium" \\
\midrule
\multicolumn{2}{c}{\textbf{Scene Description Results}} \\
\midrule
Scene Overview & Environment Type: conference room \\
& Size Estimate: approx. 20x15 feet \\
& General Description: A modern, well-lit conference room with a long table and chairs arranged for meetings. \\
\midrule
\multicolumn{2}{c}{\textbf{Key Elements}} \\
\midrule
Element 1 & Name: conference table \\
& Type: furniture \\
& Location: center of the room \\
& Description: A long rectangular table designed for meetings, surrounded by multiple chairs. \\
\midrule
Element 2 & Name: chairs \\
& Type: furniture \\
& Location: surrounding the conference table \\
& Description: Black mesh chairs that are ergonomic and designed for comfort during long meetings. \\
\midrule
Element 3 & Name: display screen \\
& Type: electronic fixture \\
& Location: adjacent to the right wall \\
& Description: A large electronic display meant for presentations or video conferencing. \\
\midrule
Element 4 & Name: window curtains \\
& Type: feature \\
& Location: along the windows on the left side \\
& Description: Blue curtains add a soft touch and natural light in the room. \\
\midrule
\multicolumn{2}{c}{\textbf{Spatial Relationships}} \\
\midrule
Relationship 1 & Reference Point: conference table \\
& Related Elements: chairs, display screen \\
& Relationship: The chairs are arranged around the table, providing seating for attendees, while the display screen is positioned at the end of the room, visible from the table. \\
\midrule
Relationship 2 & Reference Point: window \\
& Related Elements: curtains \\
& Relationship: The curtains hang directly in front of the windows, controlling the light that enters from the left side. \\
\midrule
\multicolumn{2}{c}{\textbf{Environmental Conditions}} \\
\midrule
Lighting & Well-lit with natural light from windows and overhead artificial lighting \\
\midrule
Atmosphere & Professional and inviting, suitable for meetings and discussions \\
\midrule
Notable Features & Modern design, functional layout, comfort-oriented furniture \\
\midrule
\multicolumn{2}{c}{\textbf{Navigation Guidance}} \\
\midrule
Orientation Guidance & Entering from the doorway, the conference table is the focal point. Note the display screen on the right and the windows with curtains on the left for orientation. \\
\midrule
\multicolumn{2}{c}{\textbf{System Performance}} \\
\midrule
System Response Time & 2.83 seconds \\
\end{longtable}

\onecolumn
\begin{longtable}{p{2.5cm}p{12cm}}
\caption{Detailed Analysis Results of RAG Memory Retrieval Experiment\label{tab:rag_memory_details}}\\
\toprule
\textbf{Analysis Dimension} & \textbf{Results} \\
\midrule
\multicolumn{2}{c}{\textbf{Current Scene Information}} \\
\midrule
\endfirsthead

\multicolumn{2}{c}%
{{\bfseries Table \thetable{} -- continued from previous page}} \\
\toprule
\textbf{Analysis Dimension} & \textbf{Results} \\
\midrule
\endhead

\midrule
\multicolumn{2}{r}{{Continued on next page}} \\
\bottomrule
\endfoot

\bottomrule
\endlastfoot

Current Scene Description & "The scene depicts a wooden table in a modern workspace. On the table, there is a large black box, presumably containing a product, positioned upright. Next to it is a smaller, rectangular package wrapped in gold foil with a bit of white paper peeking out. The background features light blue curtains and some office chairs in the blurred distance, suggesting a professional setting." \\
& Identified Objects: ["large black box", "small rectangular package wrapped in gold foil", "wooden table", "blue curtains", "office chairs"] \\
& Actions: ["none observed"] \\
\midrule
\multicolumn{2}{c}{\textbf{Retrieved Scenes}} \\
\midrule
Retrieved Scene 1 & Similarity: 0.7405 \\
& Scene Description: "A modern workspace featuring a wooden table. On the table, there is a large black box, likely containing a product or equipment. Beside the box sits a white paper cup with a logo, and a cable is positioned next to the cup, suggesting some electronic device may be present. The environment looks clean and organized, indicative of a professional setting." \\
& Identified Objects: ["large black box", "white paper cup", "cable", "wooden table"] \\
& Scene Type: DESCRIPTION\_SCENE \\
& Scene Analysis: [Detailed scene analysis stored in the vectorized memory system] \\
\midrule
Retrieved Scene 2 & Similarity: 0.6501 \\
& Scene Description: "The scene depicts a tabletop setting with a sleek black box labeled with branding, likely containing a headset or device. Next to it is a small white paper cup, with a wire lying across the wooden floor, suggesting a workspace or display area." \\
& Identified Objects: ["black box", "white cup", "gray wire", "wooden floor"] \\
& Scene Type: DESCRIPTION\_SCENE \\
& Scene Analysis: [Detailed scene analysis stored in the vectorized memory system] \\
\midrule
Retrieved Scene 3 & Similarity: 0.6206 \\
& Scene Description: "The scene is set in a well-lit room with large windows draped with blue curtains. There is a wooden table in the foreground. A black object, possibly an electronic device, is positioned prominently. In the background, you can see additional furniture, suggesting a workspace environment." \\
& Identified Objects: ["black electronic device", "tissue box", "coffee cup", "table"] \\
& Scene Type: DESCRIPTION\_SCENE \\
& Scene Analysis: [Detailed scene analysis stored in the vectorized memory system] \\
\midrule
\multicolumn{2}{c}{\textbf{User Interaction}} \\
\midrule
User Query & "Where is my cup?" \\
\midrule
System Response & "In the current scene, there isn't a cup specifically noted on the table. However, in previous similar scenes, a white paper cup was frequently found near the black box, often to its right side. Based on that pattern, if you continue to feel around the right side of the large black box, you might find a cup there. If you don't locate it, there may not be one in this particular meeting room, but it's always a good idea to check that area first. Let me know if you need further assistance!" \\
\end{longtable}

\end{document}